\documentclass[10pt,journal,compsoc]{IEEEtran}

\ifCLASSOPTIONcompsoc
  \usepackage[nocompress]{cite}
\else
  \usepackage{cite}
\fi

\usepackage{graphicx}
\usepackage{graphics} 
\usepackage{subfigure}
\usepackage{booktabs}
\usepackage{multirow}

\ifCLASSINFOpdf

\else

\fi

\usepackage{amsmath}
\usepackage{amssymb}
\usepackage{balance}

\usepackage{algorithm}  
\usepackage{algorithmic}

\begin{document}
\title{RGB-D SLAM in Dynamic Environments Using Point Correlations}
\author{Weichen~Dai, %
        Yu~Zhang, 
        Ping~Li, 
        Zheng~Fang,
        and~Sebastian~Scherer
\IEEEcompsocitemizethanks{
\IEEEcompsocthanksitem Weichen Dai is with the College of Control Science and Engineering, Zhejiang University, Hang Zhou, China
\IEEEcompsocthanksitem Yu Zhang and Ping Li are with the College of Control Science and Engineering and the State Key Laboratory of Industrial Control Technology, Zhejiang University, Hang Zhou, China
\IEEEcompsocthanksitem Zheng Fang is with the Faculty of Robot Science and Engineering, Northeastern University, Shenyang, China
\IEEEcompsocthanksitem Sebastian~Scherer is with the Robotics Institute, Carnegie Mellon University, Pittsburgh, PA 15213-3890 USA 
}
\thanks{The first two authors contributed equally to this work. This paper has been accepted by IEEE.} }

\IEEEtitleabstractindextext{%
\begin{abstract}

In this paper, a simultaneous localization and mapping (SLAM) method that eliminates the influence of moving objects in dynamic environments is proposed. 
This method utilizes the correlation between map points to separate points that are part of the static scene and points that are part of different moving objects into different groups.
A sparse graph is first created using Delaunay triangulation from all map points. In this graph, the vertices represent map points, and each edge represents the correlation between adjacent points.
If the relative position between two points remains consistent over time, there is correlation between them, and they are considered to be moving together rigidly. If not, they are considered to have no correlation and to be in separate groups. 
After the edges between the uncorrelated points are removed during point-correlation optimization, the remaining graph separates the map points of the moving objects from the map points of the static scene. The largest group is assumed to be the group of reliable static map points. Finally, motion estimation is performed using only these points.  
The proposed method was implemented for RGB-D sensors, evaluated with a public RGB-D benchmark, and tested in several additional challenging environments.
The experimental results demonstrate that robust and accurate performance can be achieved by the proposed SLAM method in both slightly and highly dynamic environments. Compared with other state-of-the-art methods, the proposed method can provide competitive accuracy with good real-time performance.

\end{abstract}
}

\maketitle

\IEEEdisplaynontitleabstractindextext

\IEEEpeerreviewmaketitle

\IEEEraisesectionheading{\section{Introduction}\label{sec:introduction}}

\IEEEPARstart{I}{n} recent years, vision-based motion estimation methods, including visual odometry (VO) \cite{scaramuzza2011visual} and visual simultaneous localization and mapping (vSLAM) \cite{fuentes2015visual, stachniss2016simultaneous}, have played an important role in robotic navigation thanks to the low cost and weight of cameras. 
These methods can provide six degree-of-freedom motion estimation using only input images.
However, the scenarios for which they are suitable are strictly limited by their assumptions of a static world.
In reality, the methods based on the static world assumption are influenced by or even fail because of moving objects appearing in the field of view (FOV).
A scene containing moving objects is referred to as a dynamic environment.
According to the area of the FOV that is occupied by moving objects, dynamic environments can be categorized as slightly or highly dynamic environments.
Because only a small part of the FOV is covered by moving objects in slightly dynamic environments, traditional robust estimation methods such as random sample consensus (RANSAC) \cite{fischler1981random} methods and robust weighting functions \cite{kerl2013robust, mactavish2015all} can eliminate most of the influence of moving objects.
In contrast, if a large part of the FOV is covered by moving objects, there are more observations of moving objects than observations of the static scene, which causes robust estimation methods to fail.
Therefore, traditional VO and vSLAM, which assume a static world, have limited applications in practice, and eliminating the influence of these moving objects has become an important topic.

\begin{figure}[t]
	\centering
	\includegraphics[scale=0.3]  {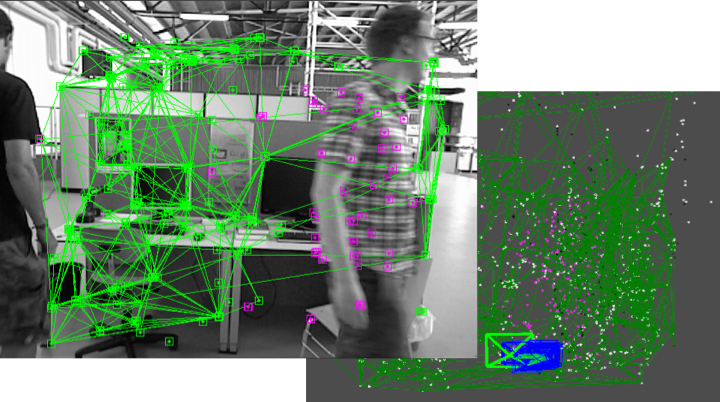}
	\caption{Results of static-point determination obtained by the proposed method. The edges between static points are shown as green lines. The features on moving objects and in the static scene are shown in pink and green, respectively, and the features in the static scene are correctly determined. Because the visual descriptor's performance is limited, a few dynamic points are inevitably matched to the features in the static scene.}
	\label{fig_exp_result}
\end{figure}

For solving the problem of motion estimation failure caused by moving objects in dynamic environments, researchers utilize multiple types of prior information such as motion consistency and semantic information. For motion consistency, the map points on the same rigid moving object have a motion consistency that is independent of the static scene. Thus, some methods \cite{lee2019robust, palazzolo2019refusion} utilize motion consistency to determine which points are on a moving object.
However, the real world also includes objects that are nonrigid owing to their dynamics and inherent deformability. Therefore, this type of method cannot determine all of the map points on the moving objects.
In addition to motion consistency, deep-learning-based methods \cite{bescos2018dynslam, runz2018maskfusion} learn semantic information from a training set as prior information to separate suspected moving objects from the static scene directly on the image. However, these methods cannot recognize unknown objects that were not present in the training set.
Moreover, their computational requirements mean that they can often struggle to run in real time in an embedded context.
Hence, it still is challenging for vSLAM and VO to provide robust and accurate navigation information in dynamic environments.

In contrast to the above two types of methods, in this study, the correlation between points is used to address the interference caused by moving objects. For simplicity, the map points on moving objects and in the static scene are called dynamic and static points, respectively. A correlation exists between static points, but there is no correlation between dynamic and static points. A segmentation method based on point correlations is proposed to exploit the connectedness of map points and separate moving objects from the static scene. The core idea of the proposed segmentation method consists of two parts. First, only the correlations between adjacent points are considered to reduce the computational requirements. In addition, through optimization of the point correlations, the edges connecting points with no correlation are removed to separate the graph. Finally, a simultaneous localization and mapping (SLAM) method that uses this segmentation method is proposed to eliminate the influence of moving objects in dynamic environments. The proposed SLAM method can accurately estimate the camera pose while eliminating the influence of slowly moving objects using the history information in previous images. The proposed SLAM method is called dynamic SLAM (DSLAM).

The main contributions of this paper are as follows: 
\begin{itemize}
	\item A segmentation method using point correlations is proposed to separate static and dynamic points. It can exploit temporal information from multiple frames to extend the captured view and is not limited to RGB-D sensors as long as the sensor can provide point-correlation measurements.
	\item A SLAM method using RGB-D sensors is proposed to improve the robustness and accuracy of motion estimation in dynamic environments. 
\end{itemize}

The rest of the paper is structured as follows. The related work is discussed in Section \ref{sec:related_work}. Section \ref{sec:statement} explains why dynamic points should be excluded from motion estimation. Section \ref{sec:segmentation} explains how to use the point correlations in detail. Section \ref{sec:SLAM_system} details the SLAM system. Section \ref{sec:expreiments} presents and discusses the experimental results, and Section \ref{sec:conclusion} presents the conclusions and plans for future work. Our experimental results can also be seen in a publicly available video.\footnotemark[1] \footnotetext[1]{youtu.be/WCOoaaVaHTw}

\section{Related Work}
\label{sec:related_work}

Vision-based motion estimation methods can be categorized as filter-based methods \cite{chiuso2002structure,davison2007monoslam} and factor graph optimization-based methods \cite{gutmann1999incremental,mouragnon2006real,klein2007parallel}. 
Because factor graph optimization is more accurate and efficient than the original approaches for SLAM based on nonlinear filtering, most advanced visual systems are based on it \cite{strasdat2012visual}.

Factor-graph-optimization-based methods can be further divided into two categories: indirect (feature-based) methods \cite{endres20133,forster2017svo,mur2017orb} and direct methods \cite{stuhmer2010real,newcombe2011dtam,steinbrucker2011real}. 
The difference between these two methods is that indirect methods use the reprojection error of the feature points, whereas direct methods directly utilize the photometric error of the raw images.
Both methods have their advantages and disadvantages.
Indirect methods can make full use of the geometric information in the correspondences between feature points and are robust to geometric noise.
Direct methods can skip the precomputation step to save computational resources. Moreover, direct methods have the ability to reconstruct a dense map.

Although the methods described above can provide excellent performance in static environments, current vision-based motion estimation methods often fail when the environment is too challenging (e.g., in highly dynamic environments) \cite{cadena2016past, saputra2018visual}. 
Existing robust estimation methods can only deal with part of the interference of moving objects in slightly dynamic environments.
Therefore, many methods have been proposed to address these problems and can be categorized into two main types: methods based on motion consistency and those based on learning.  

\textbf{(1) Methods based on motion consistency}:
Methods based on motion consistency find dynamic points on the same moving object using points with motion consistency.

Most methods treat moving objects as noise, which they filter out as soon as possible. 
Alcantarilla et al. \cite{alcantarilla2012combining} use scene flow to distinguish moving objects from the static scene, but the calculation of the scene flow is based on a result estimated by VO. Therefore, the camera pose must be estimated twice for each frame, which increases the computation time. 
Azartash et al. \cite{azartash2014visual} first partition RGB-D images. Then, the motion of each region is individually estimated to determine the regions belonging to the moving object. Their experimental results show that the accuracy slightly improves in dynamic environments. 
Zhang et al. \cite{FlowFusion2020Zhang} estimate optical flow using PWC-net and apply it to dynamic segmentation.
St{\"u}ckler and Behnke \cite{stuckler2013efficient} achieved good performance when segmenting RGB-D images into pixel regions, but this segmentation is still time-consuming.
Sun et al. \cite{sun2017improving} use the image difference and depth segmentation to filter the RGB-D data that are associated with moving objects. However, part of the depth data in the static background that is close to moving objects is also removed. 
Li and Lee \cite{li2017rgb} proposed a real-time depth edge-based RGB-D SLAM system. In the point cloud registration, each edge point has a static weight indicating the likelihood it is a part of the static background.

Some methods focus on enforcing spatial or temporal coherence among the detected dynamic points in consecutive frames.
Kim and Kim \cite{kim2016effective} obtain static regions in images by computing the depth differences between consecutive frames. However, only some of the regions belonging to moving objects have depth changes. Therefore, not all regions belonging to moving objects can be obtained.
Similarly, Jaimez et al. \cite{jaimez2017fast} introduced a joint VO and scene flow estimation method. 
Furthermore, Scona et al. \cite{scona2018staticfusion} designed a segmentation by coupling camera motion residuals, depth inconsistency, and a regularization term.

In other approaches, external sensors such as an inertial measurement unit (IMU) are leveraged to solve this problem. Kim et al. \cite{kim2015visual} combined an RGB-D camera with an IMU to estimate the camera pose. They regard the IMU information, which is relatively accurate over a short time interval, as a prior for filtering incorrect visual information from moving objects. However, the reliance on the extra IMU limits this method to scenarios with IMU sensors. 

Detecting moving objects using only RGB cameras is also a related research topic\cite{huang2020clustervo}. Most methods relying on geometric constraints leverage the properties of epipolar geometry to segment static and dynamic features. The constraints can be derived from the equation of triangulation \cite{migliore2009use} or fundamental matrix estimation \cite{kundu2009moving}.
Notably, Tan et al. \cite{tan2013robust} proposed a prior-based adaptive RANSAC algorithm to categorize points according to reprojection error. Then, the hypothesis that the texture of the static scene is evenly distributed is used to determine static points. However, this hypothesis may fail in situations where most of the FOV is covered by moving objects. Another approach also leverages the reprojection error. Zou et al. \cite{zou2012coslam} use multiple cameras that can capture a broader view to avoid occlusion. To benefit from this additional information, they employ intercamera pose estimation and intercamera mapping to deal with dynamic objects and enhance the system's robustness. In the proposed method, point correlation is employed to segment the static and dynamic points. In contrast to the reprojection error, which only considers the correlation between frame and map points, the point correlation further determines the correlation between map points. Therefore, in contrast to the expansion of the captured view using multiple cameras, the method proposed in this paper expands the captured view using multiple temporal frames.

The methods described above provide good performance in dynamic environments. However, it is difficult for them to maintain robust motion estimation when a slowly moving object enters the FOV. Moreover, most methods can only be applied to a specific sensor.

\textbf{(2) Methods based on learning}:
Other methods, such as \cite{redmon2016you, he2018mask, dai2016instance}, use pretrained learning-based methods to detect potential moving objects. Most methods of this type treat dynamic region features as outliers or exclude those regions from the image.

Kitt et al. \cite{kitt2010moving} classify feature points to distinguish dynamic and static points. However, the classifier has to be trained in advance, which prevents this method from being used to explore unknown environments. 
Riazuelo et al. \cite{riazuelo2017semantic} used semantic information to partition the image regions of people to eliminate the influence of walking people. 
B{\^a}rsan et al. \cite{barsan2018robust} also used an instance-aware semantic segmentation algorithm to recognize moving objects from a single image.
Bescos et al. \cite{bescos2018dynslam} proposed DynaSLAM, which combines a Mask R-CNN \cite{he2017mask} prior with multi-view geometry to partition an image and determine which regions in the image belong to moving objects.
Moreover, some other methods additionally track moving objects to provide a complete three-dimensional (3D) map and the trajectory of these moving objects.
Qiu et al. \cite{qiu2019tracking} eliminate the interference of moving objects using visual-inertial sensing and use Re-3 \cite{gordon2018re} to track moving objects.
Yang et al. \cite{yang2019cubeslam} utilize moving objects and motion model constraints to improve the camera pose estimation.
Strecke et al. \cite{strecke2019fusion} initially detected and segmented new movable instances using Mask R-CNN.
R{\"u}nz and Agapito \cite{runz2017co} partition an image using semantic cues to improve the accuracy of segmentation. However, only a few 3D models of different objects are maintained, and the method only works at a low frame rate.
To robustly estimate motion, Xu \cite{xu2019mid} proposed a system to generate an object-level dynamic volumetric map from a single RGB-D camera.

These approaches can work well in environments that only contain the types of objects on which the classifier was trained, but they can fail in the presence of unknown objects that are not present in the training set. 
Furthermore, state-of-the-art segmentation methods such as Mask R-CNN are still computationally intensive and often have to be run in a background thread in practice to amortize their cost over several frames \cite{runz2018maskfusion}. 

In contrast to the above methods, we propose a method based on point correlations.  Instead of preprocessing the image as in image segmentation methods, the proposed method relies on the fact that there are correlations between two static points but not between dynamic and static points. 
Because the point correlations can be obtained irrespective of the type of sensors used, the proposed method has the potential to be applied to various types of sensors.

\section{Problem Statement}

\label{sec:statement}

\begin{figure}[t]
	\centering
	\includegraphics[scale=0.15]{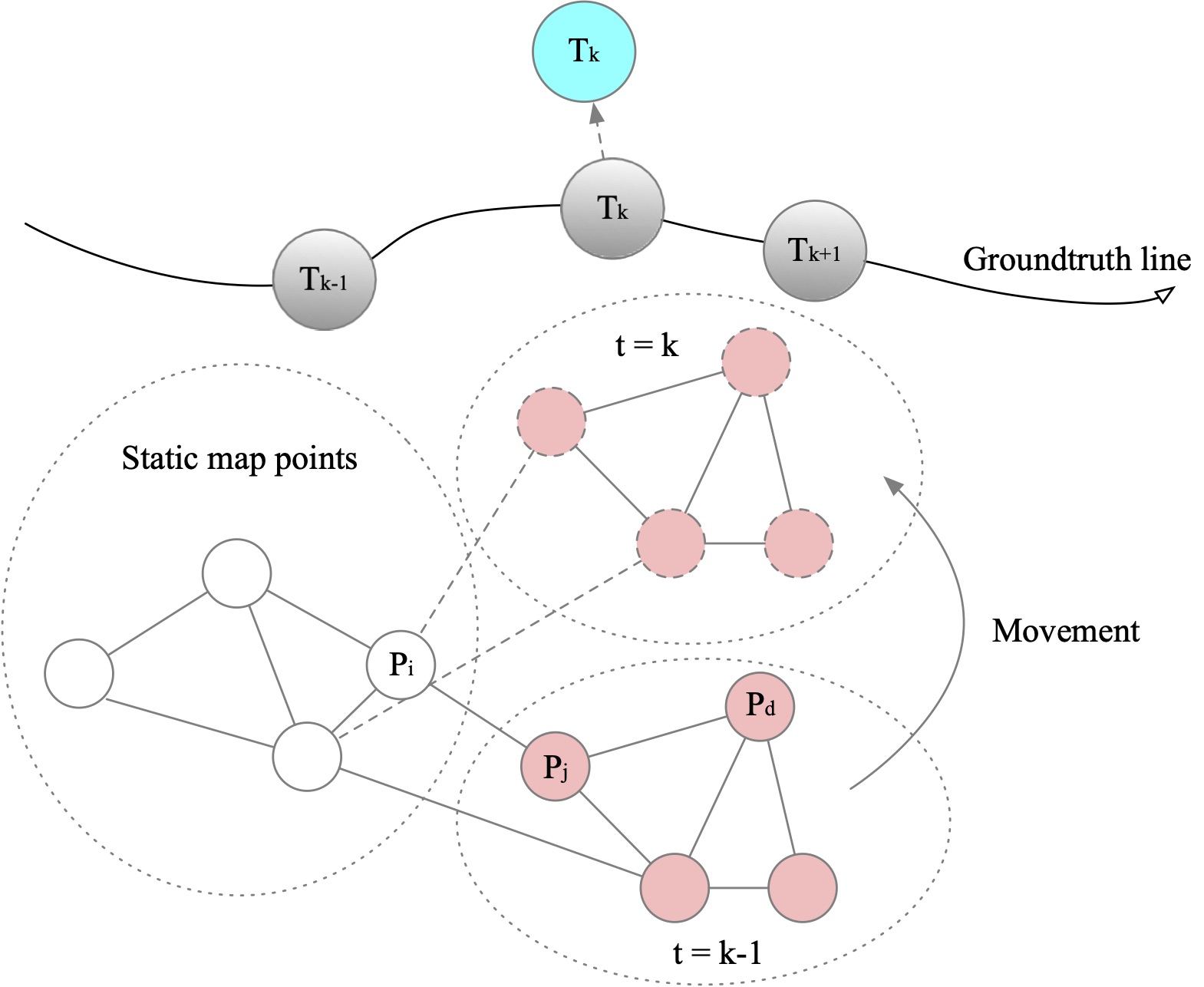}
	\caption{Illustration of the negative effects caused by dynamic points. The gray circles are the correct results from pose estimation. The white circles are static map points, and the pink circles are points on a moving object. The movement of dynamic points destroys the consistency of motion estimation, as shown in the figure, where the unrecognized dynamic points influence the estimation of $\mathbf{T}_k$, so it moves from the correct estimation result (gray circle) to the incorrect result (blue circle). Besides, the movement of the dynamic points also destroys the point correlations between points $\mathbf{p}_i$ and $\mathbf{p}_j$.}
	\label{fig:dynamic}
\end{figure}

In this section, we briefly introduce motion estimation based on the static world assumption and explain why estimation is influenced by moving objects. 

Bundle adjustment \cite{barfoot2017state} under the static world assumption is used in most VO and vSLAM methods. As shown in Figure \ref{fig:dynamic}, the states $\mathbf{T}_k$ and $\mathbf{p}_i$ are estimated using bundle adjustment. Here, $\mathbf{T}_k$ is the transformation matrix representing the pose of the sensor at time $k$, and $\mathbf{p}_i$ is the $i$-th coordinate representing the position of the $i$-th map point, where $k =1, \ldots, K$ and $i = 1,\ldots, M$. All states are represented as
\begin{equation}
\mathbf{x} = \{\mathbf{T}_1, ..., \mathbf{T}_K, \mathbf{p}_1, ... , \mathbf{p}_M \},
\end{equation}
where $\mathbf{x}_{ik} = \{\mathbf{T}_k,\mathbf{p}_i\}$ is the subset of states, including the $k$-th pose and $i$-th map point.
Regardless of the type of visual sensors used, the measurement $\mathbf{y}_{ik}$ corresponding to the observation of point $i$ from pose $k$ can be expressed as
\begin{equation}
\mathbf{y}_{ik} = \mathbf{g}(\mathbf{x}_{ik}) + \mathbf{n}_{ik},
\label{equ:static_obs}
\end{equation}
where $\mathbf{g}(\mathbf{x}_{ik})=\mathbf{s}(\mathbf{r}(\mathbf{x}_{ik}))$ is the measurement model,  $\mathbf{r}(\mathbf{x}_{ik})={\mathbf{T}_k \cdot \overline{\mathbf{p}}_i}$, $\mathbf{s}(\cdot)$ is the nonlinear sensor model producing observations from the point in the frame (e.g., the camera model), the
notation $\overline{\mathbf{p}}$ makes an augmented vector from a 3D point $\mathbf{p} \in \mathbb{R}^3$ to homogeneous coordinates, and $\mathbf{n}_{ik} \sim \mathcal{N} (\boldsymbol{\mu}_{ik},\mathbf{C}_{ik}) $ is additive Gaussian noise, where $\boldsymbol{\mu}_{ik} = \mathbf(0)$ and $\mathbf{C}_{ik}$ denotes the Gaussian noise covariance associated with $\mathbf{y}_{ik}$.

Typically, a maximum likelihood approach is used. This approach finds the optimal states $\mathbf{x}$ that maximize the probability of obtaining the actual measurements, $\mathbf{x}^* = \operatorname*{argmax}_\mathbf{x} P(\mathbf{y}|\mathbf{x})$, where $\mathbf{y}$ represents all measurements.
Traditional methods neglect the correlations between points, so that they can perform the estimation in real time, and the objective function is hence written as
\begin{equation}
{J}_{ba}(\mathbf{x}) =   \frac{1}{2} \sum_{i,k} {\mathbf{e}_{y,ik}(\mathbf{x})^T \mathbf{C}^{-1}_{ik} \mathbf{e}_{y,ik}(\mathbf{x})},
\label{equ:opti_trad}
\end{equation}
where $\mathbf{x}$ is the full state that we wish to estimate and
\begin{equation}
\mathbf{e}_{y,ik}(\mathbf{x}) = \mathbf{y}_{ik}-\mathbf{g}(\mathbf{x}_{ik}).
\end{equation}

The usual approach to this estimation problem is to apply the Gauss--Newton method.
In the Gauss--Newton method, the Hessian structure of the objective function in Eq. (\ref{equ:opti_trad}) is
\begin{equation}
\mathbf{H} = 
\begin{bmatrix}
\mathbf{H}_{pp} & \mathbf{H}_{pg}  \\ \mathbf{H}_{gp} & \mathbf{H}_{gg} 
\end{bmatrix},
\end{equation}
where $\mathbf{H}_{pp}$, $\mathbf{H}_{pg}$, and $\mathbf{H}_{gg}$ denote the pose--pose, pose--geometry, geometry--geometry blocks, respectively.
Here, ``pose'' refers to the pose parameters of all of the camera poses in $\mathbf{x}$, and ``geometry'' refers to the geometry parameters of all of the map points in $\mathbf{x}$. 
Because $\mathbf{H}_{gg}$ is diagonal, methods such as Cholesky factorization and the Schur complement method \cite{grisetti2011g2o, polok2013incremental} can be used to solve it efficiently.

The above descriptions all assume a static world.
For dynamic environments, the dynamic points do not satisfy the measurement model in Eq. (\ref{equ:static_obs}). The measurement model of a dynamic point $\mathbf{p}_d$ should be 
\begin{equation}
\mathbf{y}_{dk} = \mathbf{g}(\mathbf{x}_{dk}+\mathbf{\bar{v}}_{dk}T) + \mathbf{n}_{dk},
\label{equ:dynamic_meas}
\end{equation}
where $\mathbf{\bar{v}}_{dk}$ is the average velocity of the moving object over the interval $T$ between time stamps $k-1$ and $k$.
If the measurement model in Eq. (\ref{equ:static_obs}) is applied to the dynamic points, $\mathbf{\bar{v}}_{dk}T $ cannot be modeled and will generate extra noise, thereby jeopardizing the estimation result. Some robust methods, such as RANSAC, can only be used to reduce the influence of moving objects in slightly dynamic environments. However, robust estimation methods will also fail if there are moving objects with rich texture occupying the majority of the FOV.
Therefore, errors will be introduced into the result if methods based on the static world assumption are used for dynamic environments. The estimation could even fail.

In dynamic environments, motion estimation methods should distinguish between static and dynamic points and then apply the measurement models in Eqs. \eqref{equ:static_obs} and \eqref{equ:dynamic_meas} separately. In reality, the velocities of moving objects cannot be obtained without external sensors. Because the velocities are unknown, only static points using the model in Eq. \eqref{equ:static_obs} should be used to acquire the result. 
Therefore, the static points must be accurately determined so that the map points used for motion estimation do not contain dynamic points. A segmentation method for determining the static points is proposed in the next section.

\section{ Segmentation Using Point Correlations}
\label{sec:segmentation}

\begin{figure}[t]
	\centering
	\includegraphics[scale=0.15]{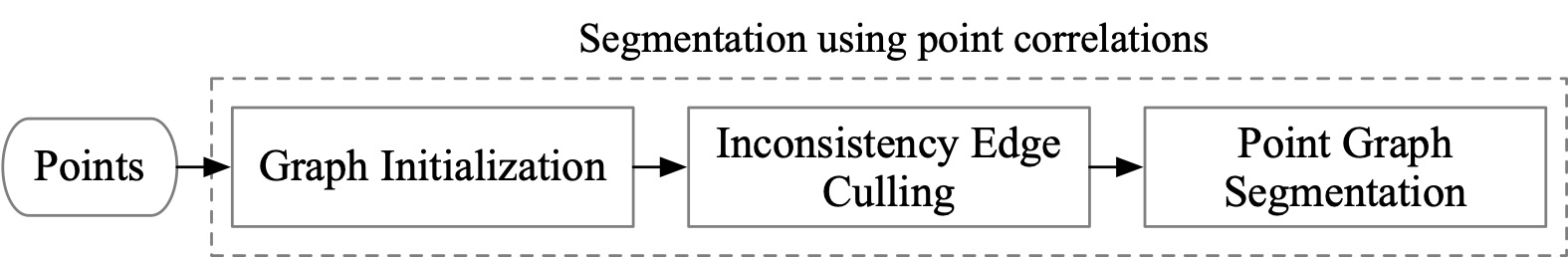}
	\caption{Overview of the proposed segmentation method, which comprises
	three steps to divide the point cloud into different components with different motion patterns: graph initialization, inconsistent edge culling, and point graph segmentation.}

	\label{fig:steps}
\end{figure}

\begin{figure}[bpt]
	\centering
	\begin{minipage}[t]{0.15\textwidth}
		\centering
		\includegraphics[width=\textwidth]  {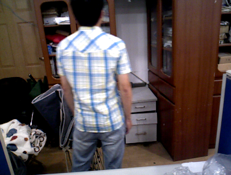}
		(a)
	\end{minipage}
	\begin{minipage}[t]{0.15\textwidth}
		\centering
		\includegraphics[width=\textwidth]  {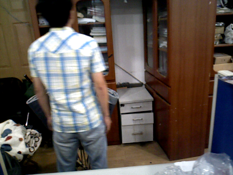}
		(b)
	\end{minipage}
	\begin{minipage}[t]{0.15\textwidth}
		\centering
		\includegraphics[width=\textwidth]  {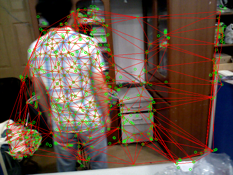}
		(c)
	\end{minipage}
	\begin{minipage}[t]{0.15\textwidth}
		\centering
		\includegraphics[width=\textwidth]  {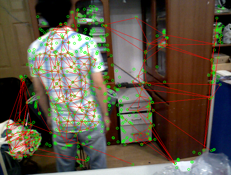}
		(d)
	\end{minipage}
	\begin{minipage}[t]{0.15\textwidth}
		\centering
		\includegraphics[width=\textwidth]  {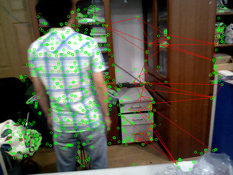}
		(e)
	\end{minipage}
	\begin{minipage}[t]{0.15\textwidth}
		\centering
		\includegraphics[width=\textwidth]  {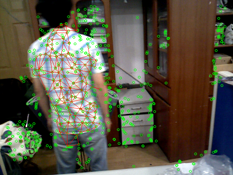}
		(f)
	\end{minipage}
	\caption{Two-dimensional example of the segmentation method using point correlations: (a) current frame; (b) reference frame; (c) structure graph of the matched feature points in the reference frame created by Delaunay triangulation; (d) graph after removing inconsistent edges; (e) extracted largest region, which belongs to the static scene; and (f) moving object region separated from the static scene.}
	
	\label{fig:seg_demo}
\end{figure}

The proposed segmentation method is based on point correlations.
Because the relative positions between static points do not change with time whereas the relative positions between static and dynamic points change with time.

The proposed segmentation method can be divided into three steps, as shown in Figure \ref{fig:steps}.
In the first step, we build a graph to represent the correlations between adjacent map points. The second step is to optimize the geometric parameters of all map points and remove inconsistent edges to partition the graph. In the third step, a depth-first search (DFS) is used to determine connected components. The points in the same component have consistent motion, and the points in different components have independent motion patterns. Figure \ref{fig:seg_demo} shows a 2D example of two frames in which the static and dynamic point groups are separated.

\subsection{Step 1: Graph Initialization}
\label{sec:segmentation_init}

In this step, a graph is built using map points to represent point correlations.
Each edge of the graph $\mathcal{G}$ represents a possible correlation between two connected points.
The edge between points $\mathbf{p}_i$ and $\mathbf{p}_j$ is indexed as a vector $\mathbf{l}_{ij}$, defined as follows: 
\begin{equation}
\mathbf{l}_{ij} = \mathbf{p}_{i} - \mathbf{p}_{j}.
\end{equation}
Next, Delaunay triangulation \cite{barber1996quickhull} is applied to reduce the complexity of the construction of the 3D point graph.
In Delaunay triangulation, for a given set of discrete points, triangulation is carried out such that no point is inside the circumcircle of any triangle. It maximizes the minimum angle of all of the angles of the triangles. Therefore, only adjacent feature points are connected in the graph. The generated graph structure is similar to a sparse mesh, as shown in Figure \ref{fig:seg_demo}(c). In other words, because of its sparsity, only the correlations between adjacent connected points are verified in the next step.

\subsection{Step 2: Inconsistent Edge Culling}

The initial graph $\mathcal{G}$ obtained from Step 1 represents the correlation between adjacent map points.
At time stamp $k$, the point-correlation measurement $\mathbf{z}_{ijk} $ of $\mathbf{l}_{ijk}$ has the form
\begin{equation}
\mathbf{z}_{ijk} = \mathbf{y}_{ik} - \mathbf{y}_{jk}= \mathbf{h}(\mathbf{l}_{ijk}) + \mathbf{n}_{ijk},
\label{equ:pcmm}
\end{equation}
where 
$\mathbf{l}_{ijk}$ is the edge between $\mathbf{p}_{i}$ and $\mathbf{p}_{j}$ at time stamp $k$,
$\mathbf{n}_{ijk}$ is additive Gaussian noise, and $\mathbf{h}()$ is the observation model of the edge, derived as
\begin{equation}
\mathbf{h}(\mathbf{l}_{ijk}) = \mathbf{s}(\mathbf{r}(\mathbf{x}_{ik})) - \mathbf{s} (\mathbf{r}(\mathbf{x}_{jk})) 
.
\end{equation}
The set of measurements collected up to time $k$ is
\begin{equation}
	\mathbf{z}_{k} = \{\mathbf{z}_{ijk}\}_{(ij) \in \mathcal{G}}.
\end{equation}
Note that we use the simpler notation 
\begin{equation}
	\mathbf{z} = \{\mathbf{z}_{0},...,\mathbf{z}_{k}\}
\end{equation}
to express all of the measurements that are available.
The set of all the geometric parameters of the map points is denoted as $\mathbf{x}_g$, which is a subset of $\mathbf{x}$. 
Similar to bundle adjustment, we also set up the estimation of $\mathbf{x}_g$ using the maximum likelihood framework.
The objective function of point-correlation optimization is defined as
\begin{equation}
{J}_p(\mathbf{x}_g) = \frac{1}{2} \sum_{ij,k} { \mathbf{e}_{z,ijk}(\mathbf{l}_{ijk})^T \mathbf{C}^{-1}_{ijk} \mathbf{e}_{z,ijk}(\mathbf{l}_{ijk}) },
\label{equ:geo_prior}
\end{equation}
where $\mathbf{C}_{ijk}$ is the covariance matrix associated with the $ijk$-th measurement and $\mathbf{e}_{z,ijk}$ is the error term and is defined as
\begin{equation}
\label{equ:opt_points}
\mathbf{e}_{z,ijk}(\mathbf{x}_g) = \mathbf{z}_{ijk}-\mathbf{h}(\mathbf{l}_{ijk}).
\end{equation}
Combining the objective function of the bundle adjustment problem in Eq. (\ref{equ:opti_trad}), the final objective function is derived as
\begin{equation}
{J}(\mathbf{x}) = {J}_p(\mathbf{x}_g) + {J}_{ba}(\mathbf{x}).
\label{equ:full}
\end{equation}

As shown in Figure \ref{fig:dynamic}, the estimated pose at time $k$ is disturbed by interference from a moving object. 
Because the relative positions between static and dynamic points have changed, the edges between these points no longer have consistent observations in multiple frames. 
These edges are called inconsistent edges, meaning that there are no correlations between connected map points.
In the process of optimizing $x_p$, the squared Mahalanobis length of the error is used to determine whether a measurement is an outlier, and a chi-squared value is chosen as the threshold according to the P-value. 
the measurements of inconsistent edges cannot fit the point-correlation model and are removed as outliers in the iterations.
Therefore, if all observations of an edge are outliers after optimization, this edge is identified as an inconsistent edge and removed from $\mathcal{G}$.
Because there are no correlations between any dynamic and static points, the remaining graph $\mathcal{G}$ corresponds to the segmentation result.

\begin{figure}[t]
	\centering
	\includegraphics[scale=0.04]{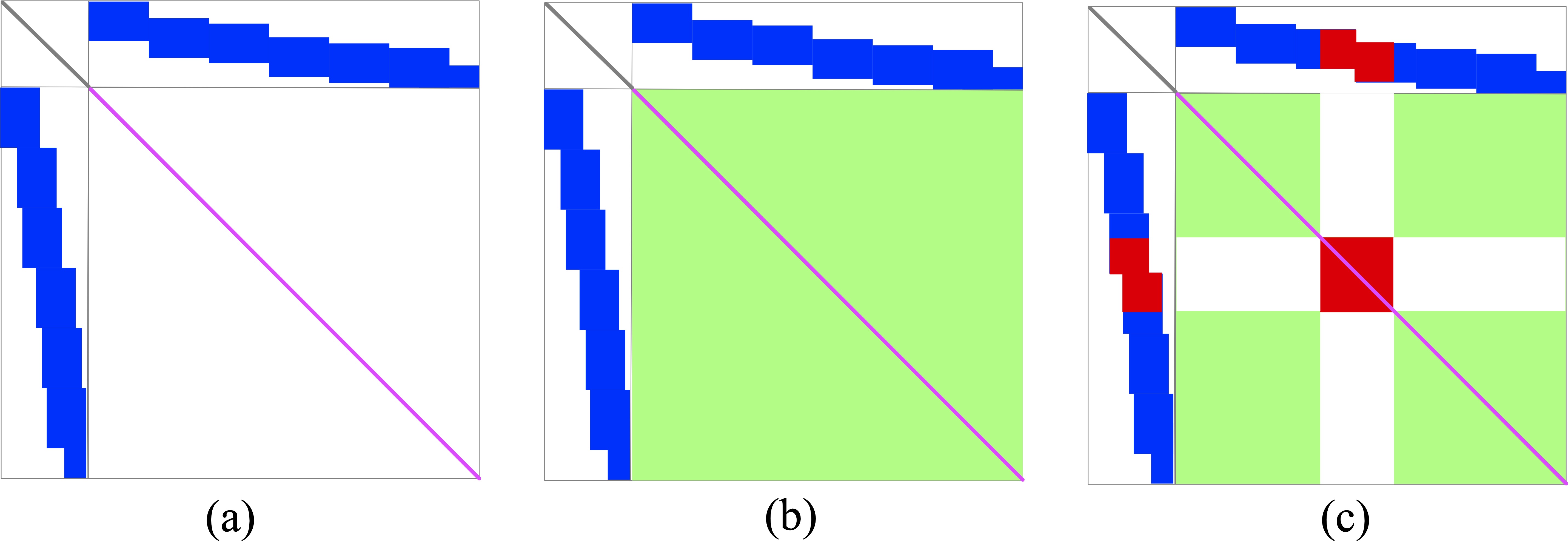}
	\caption{Example of a Hessian structure.
		(a) Hessian structure of bundle adjustment in Eq. (\ref{equ:opti_trad}). 
		Because the geometry--geometry block is diagonal, the Hessian can be efficiently solved using the Schur complement.
		(b) Hessian structure of the objective function in Eq. (\ref{equ:full}).
		The geometry--geometry block of the Hessian structure is the green block with the pink diagonal.
		(c) The result after the inconsistent edge observations are removed from the Hessian structure. The green blocks indicate the point correlations, and the blue blocks indicate the geometry--pose correlations.
		The red blocks indicate moving objects in the camera image.
		Therefore, if inconsistent measurements in the point correlations can be determined, the two point clusters with different motion consistencies are separated.
}
\label{fig:hassian_change}
\end{figure}

The process of culling the inconsistent edges can also be shown as changes in the Hessian structure of the objective function in Eq. (\ref{equ:full}).
As shown in Figure \ref{fig:hassian_change}, the geometry--geometry block on the Hessian side is divided into multiple independent blocks after the outlier observations corresponding to the inconsistent edges have been removed.
The points in the blocks do not correlate with points in the other blocks. From the example shown in Figure \ref{fig:hassian_change}(c), two point groups that have their own motion patterns have been separated.
However, objective function ${J}(\mathbf{x})$ in Eq. (\ref{equ:full}) destroys the specific sparsity pattern of the Hessian \cite{grisetti2011g2o,polok2013incremental} of the objective function in Eq. (\ref{equ:opti_trad}).
The introduction of correlations between the geometric parameters makes joint optimization real-time infeasible \cite{engel2017direct}, which depends on the block-diagonal geometry--geometry block.
For retaining the efficiency of solving the bundle adjustment problem, the optimization of the objective function in Eq. (\ref{equ:geo_prior}) must be performed separately from bundle adjustment. 
In addition, the Huber norm, as used in bundle adjustment, is also applied to the edge residuals to improve robustness.

The 2D example shown in Figure \ref{fig:seg_demo} illustrates this process.
After the camera position changes, the distances between static points do not significantly change, whereas the distances between static and dynamic points change by large amounts. Therefore, the edges between static and dynamic points are removed. 
The remaining connected components of the graph describe the separated moving objects and static scene. 
As a result, the person shown in Figure \ref{fig:seg_demo}(f) is successfully separated from the static background.

\subsection{Step 3: Determination of Static Points}

After removing the inconsistent edges, 
search algorithms such as the DFS algorithm are used to check whether $\mathcal{G}$ is divided into several isolated subgraphs (connected components).
If there are different connected components, these components represent different point groups with different motion patterns.

For determining which points are reliable static points, the connected component with the largest volume is assumed to consist of reliable static points for two reasons: 
\begin{itemize}
\item In a map built over a period of time, the spatial volume of the static points, whose volumes grow the most over time, is generally the largest.
\item Static points are generally evenly distributed in 3D space, whereas dynamic points are generally only distributed on a surface because the camera can only observe one side of the surface of a moving object. Therefore, the volume of the dynamic points will be much smaller than that of the static points.
\end{itemize}
For these reasons, the points of the connected component with the largest volume are identified as the reliable static points. In the subsequent calculations, only reliable static points are used for motion estimation.

\section{DSLAM in Dynamic Environments Using RGB-D Sensors}

\label{sec:SLAM_system}

\begin{figure*}[t]
	\centering
	\includegraphics[scale=0.15]{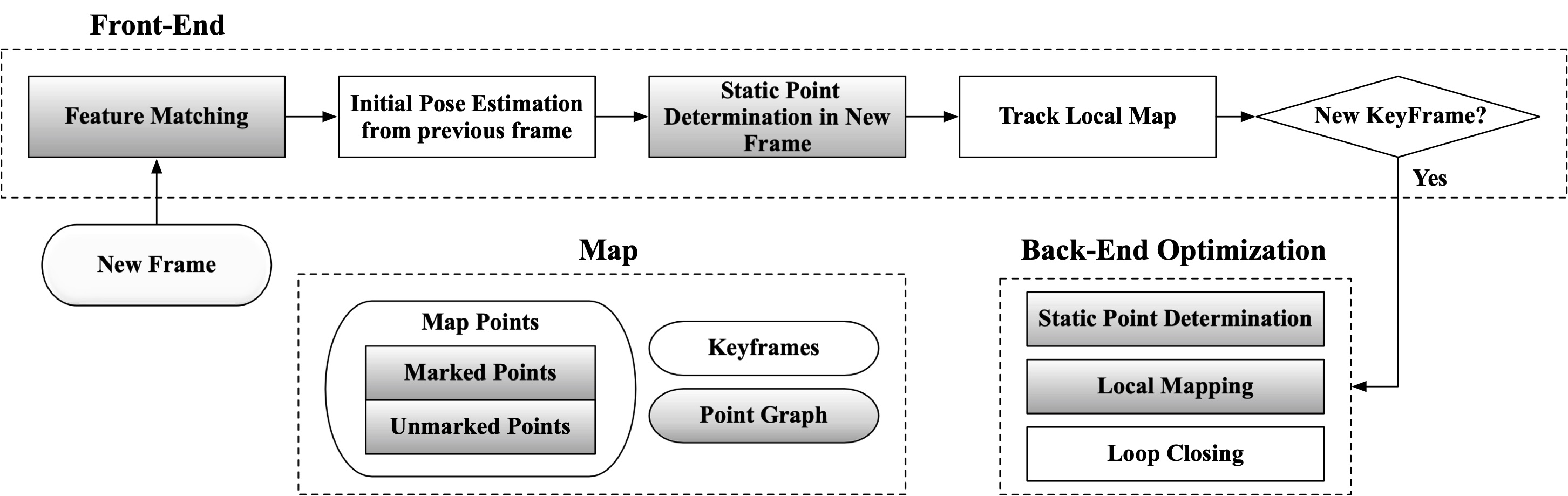}
	\caption{System overview. Static point determination is performed in both the front and back ends. Before tracking the local map, the dynamic points are removed from the estimation data. In the back end, the determination of static points and local mapping simultaneously occur. The shaded rectangles indicate components specifically modified for dynamic environments.}

	\label{fig:overview}
\end{figure*}

A SLAM method that integrates the segmentation method proposed in Section \ref{sec:segmentation} called dynamic SLAM (DSLAM) is proposed for dynamic environments. This SLAM method is implemented on RGB-D sensors, which provide color images with a synchronized depth image. DSLAM is built on ORB-SLAM2 \cite{mur2017orb} and consists of front and back ends. The entire pipeline is illustrated in Figure \ref{fig:overview}, where each component modified for dynamic environments is shown in gray. As shown in Figure \ref{fig:overview}, the static point determination module implementing the proposed segmentation method is added to eliminate the influence of moving objects. 
Moreover, the map points are divided into marked and unmarked points. The marked points indicate that a map point is a reliable static label. The unmarked points indicate that a map point cannot be reliably identified as a static point. Note that all new map points are initialized as marked points and are checked by the static point determination module later.

The front end is responsible for estimating the motion between consecutive frames and determining when to insert a new keyframe.
Therefore, the front-end tasks consist of motion estimation, determination of the static points in the current frame, and determination of new keyframes.
Because RGB-D sensors can capture 3D information, the initialization of the system is finished when the first frame has created a sufficient number of available map points.

After initialization is complete, the specific steps are as follows. 
For each new frame, initial feature matching is performed with the marked points that were successfully tracked in the previous frame. Then, in the initial pose estimation step, a pose is estimated using motion-only bundle adjustment with all matches. This step excludes the incorrect feature matching results. In the initial pose estimation step, note that the features of fast-moving objects are rarely matched with the existing map points. Meanwhile, this step can also exclude some of the correspondences on fast-moving objects with significant reprojection error. Therefore, even if the initial pose is obtained before the static point determination, only a few features related to moving objects may jeopardize motion-only bundle adjustment.
If the tracking is not lost, the static point determination step is performed for the tracked map points. Afterward, the matches between the untracked features and the local map points are searched for by reprojection and the pose is optimized again in the track local map step. Finally, the tracking thread determines whether a new keyframe needs to be inserted in the new keyframe decision step.

The back-end tasks consist of optimal static scene reconstruction, determination of the static points using sliding windows, and pose graph optimization.
These tasks are performed by three modules running in three independent threads: the local mapping, static point determination, and loop closing modules.
The local mapping module processes each new keyframe and uses the local bundle adjustment to optimize all marked points.
The static point determination module uses point-correlation optimization to determine reliable static points and remove the marked status from the other points to eliminate the influence of moving objects.
The loop closing module searches for loop closures using the bag-of-words method. If a new loop closure is found, pose graph optimization is performed to eliminate the accumulated global drift and achieve global consistency.

In the next sections, changes made to three components in DSLAM for dynamic environments are introduced: the determination of static points, feature matching, and map management.

\subsection{Determination of Static Points for RGB-D Sensors}
This section introduces the measurement models related to RGB-D sensors. Then, the static point determination modules in the front and back ends are described in detail. Both implementations utilize graph optimization to optimize the objective function Eq. (\ref{equ:geo_prior}) using nonlinear least-squares techniques \cite{grisetti2011g2o}.

\subsubsection{Formulation of the Point-Correlation Measurement Model for RGB-D Sensors}
The measurement models related to RGB-D sensors should be described before describing the proposed segmentation method.
Hence, the map-point and point-correlation measurement models are  described in this section.

\begin{figure}[bpt]
	\centering
	\begin{minipage}[t]{0.15\textwidth}
		\centering
		\includegraphics[width=\textwidth]  {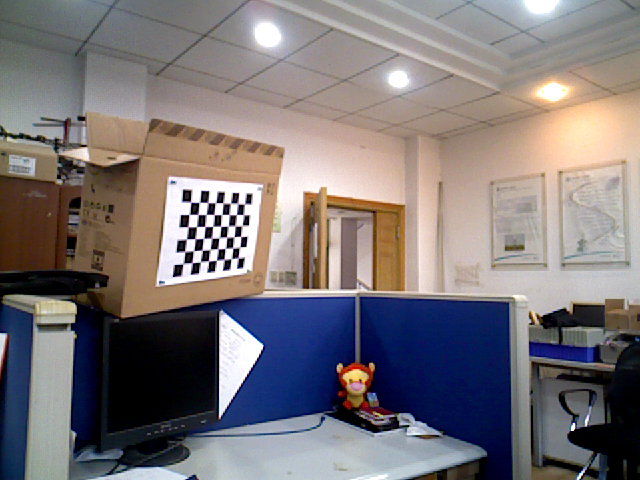}
		(a)  RGB
	\end{minipage}
	\begin{minipage}[t]{0.15\textwidth}
		\centering
		\includegraphics[width=\textwidth]  {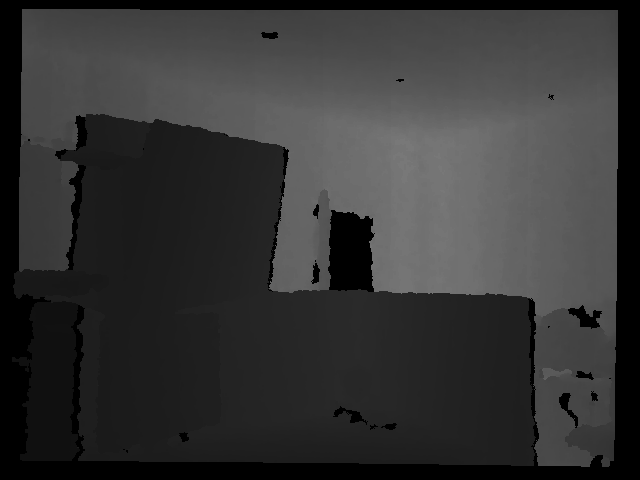}
		(b)  Depth
	\end{minipage}
	\begin{minipage}[t]{0.15\textwidth}
		\centering
		\includegraphics[width=\textwidth]  {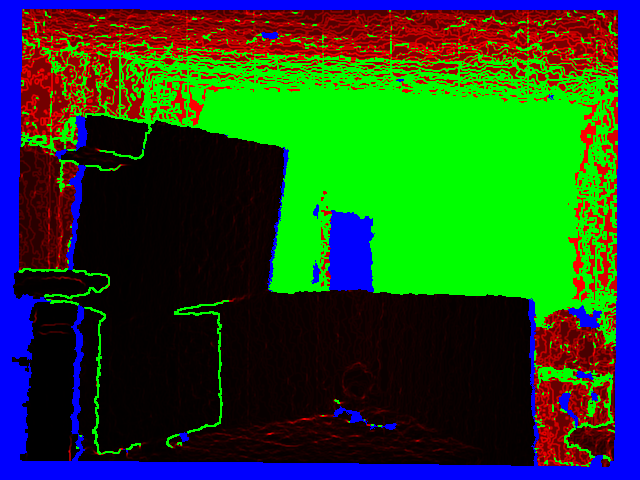}
		(c)  Uncertainty
	\end{minipage}
	\caption{Images captured by an ASUS Xtion Pro camera and the uncertainty results for depth: (a) color image, (b) depth image registered with the color image, and (c) the uncertainty of the depth image. Blue pixels indicate no data because the depth exceeds the range of the depth camera. Green pixels indicate that the uncertainty of the measured depth is too large to be trusted. Colors from black to red denote the uncertainties of the pixels with valid depths (black = low uncertainty, red = high uncertainty).
	}
	\label{fig:uncertainty}
\end{figure}

\textbf{Map-point measurement:}
For RGB-D sensors, $\mathbf{y}_{ik}$ can be specified as 
\begin{equation}
\label{equ:rgbd_yik}
\mathbf{y}_{ik} 
=
\begin{bmatrix}
(u - c_u) \times d/f_u \\
(v - c_v) \times d/f_v \\
d 
\end{bmatrix} 
= \mathbf{g}(\mathbf{x}_{ik}) + \mathbf{n}_{ik},
\end{equation}
where ($c_u$, $c_v$) is the camera's principal point and ($f_u$, $f_v$) is the focal length. It is assumed that the measurement of map point $\mathbf{p}_{i}$ is $(u,v)$ in the RGB image and its corresponding depth measurement is $d$. Moreover, $\mathbf{g}(\mathbf{x}_{ik})$ is the RGB-D sensor model, expressed as 
\begin{equation}\mathbf{g}(\mathbf{x}_{ik})=\mathbf{s}_{RGB-D}(\mathbf{T}_k \cdot \overline{\mathbf{p}}_i),
\end{equation}
where $\mathbf{s}_{RGB-D}(\mathbf{\rho}) = \left[ 
\begin{matrix} 
1 & 0 & 0 & 0 \\ 0 & 1 & 0 & 0 \\ 0 & 0 & 1 & 0
\end{matrix} 
\right] \mathbf{\rho}$, and $\mathbf{\rho}$ is a 4D vector.
For convenience, $[a,b,c]^T$ is used to represent $\mathbf{y}_{ik}$, i.e., 
\begin{equation}
\mathbf{y}_{ik} 
=
\begin{bmatrix}
a \\
b \\
c 
\end{bmatrix}.
\end{equation}

In addition to $\mathbf{g}(\cdot)$, the uncertainty of the noise should also be determined.
Given the particular characteristics of the structured light technology used by RGB-D sensors, the depth measurement error dramatically increases with the sensing depth. Therefore, the uncertainty in the depth measurement should be modeled.
Khoshelham \cite{khoshelham2012accuracy} modeled the depth measurement of Kinect-style devices and concluded that the uncertainty in each depth measurement is proportional to the square of its depth, i.e., 
\begin{equation}
\sigma_d= \frac{1}{f_v}\sigma d^2,
\end{equation}
where $\sigma$ and $\sigma_d$ are respectively the standard deviation of the measured normalized disparity and the standard deviation of the calculated depth.
In this model, $\sigma_d$ increases as $d$ increases.
However, this model does not consider the problem of more significant uncertainty at object edges. 
For establishing a more accurate uncertainty model, a Gaussian mixture model is used \cite{dryanovski2013fast}. In this model, $u$ and $v$ are assumed to be independent random variables distributed according to the normal distributions $\mathcal{N}(u,\sigma_u)$ and $\mathcal{N}(v,\sigma_v)$, respectively. 
The random variable $c$ is defined as a mixture of the $d$ variables in a local window $ {i\in [u-1,u+1], j\in[v-1,v+1]}$. The mean and variance of the resulting Gaussian mixture are 
\begin{equation}
\begin{aligned}
\hat\mu_c &=\sum_{\mathbf{q} \in \mathcal{N}_{uv} }w_{\mathbf{q}}(d_{\mathbf{q}}), \\
\hat\sigma_c &=\sum_{\mathbf{q} \in \mathcal{N}_{uv} }w_{\mathbf{q}}(\sigma_{d_{\mathbf{q}}}^2 + d_{\mathbf{q}}^2)-\hat\mu_c^2,
\end{aligned}
\label{equ:mix_gauss}
\end{equation}
where $\mathcal{N}_{uv}$ is the position set located in the window with $(u,v)$ as the center, $d_{\mathbf{q}}$ is the depth measurement at position ${\mathbf{q}}$, and the weight $w_{\mathbf{q}}$ is chosen according to kernel $W$, which is expressed as follows: 
\begin{equation}
W = \frac{1}{16} \left[ 
\begin{matrix} 
1 & 2 & 1 \\ 2 & 4 & 2 \\ 1 & 2 & 1
\end{matrix} 
\right].
\end{equation}
An example is shown in Figure \ref{fig:uncertainty}(c).

Therefore, based on Eq. \eqref{equ:mix_gauss}, the 3D covariance $\mathbf{\Sigma}_{ik}$ of the measurement of the $i$-th map point at time $k$ is assigned using 
\begin{equation}
\mathbf{C}_{ik} =  \left[ 
\begin{matrix} 
\sigma_a^2 & \sigma_{ab} & \sigma_{ac} \\ \sigma_{ba} & \sigma_b^2 & \sigma_{bc} \\ \sigma_{ca} & \sigma_{cb} & \sigma_{c}^2
\end{matrix} 
\right],
\end{equation}
where
\begin{equation}
\begin{aligned}
\sigma_a^2 &= \frac{\hat\sigma_{c}^2(u-c_u)(v-c_v)+\sigma_u^2(\hat\mu_c^2+\hat\sigma_{c}^2)}{f_x^2}, \\
\sigma_b^2 &= \frac{\hat\sigma_{c}^2(u-c_u)(v-c_v)+\sigma_v^2(\hat\mu_c^2+\hat\sigma_{c}^2)}{f_y^2}, \\
\sigma_{ac} &= \sigma_{ca}  = \hat\sigma_{c}^2\frac{u-c_u}{f_u}, \\
\sigma_{bc} &= \sigma_{cb} = \hat\sigma_{c}^2\frac{v-c_v}{f_v}, \\
\sigma_{ab} &= \sigma_{ba} =  \hat\sigma_{c}^2\frac{(u-c_u) (v-c_v)}{f_u f_v}, \\
\sigma_{c}^2 &=  \hat\sigma_{c}^2.
\end{aligned}
\end{equation}

\textbf{Point-correlation measurement model:}
On the basis of the point measurement model in Eq. \eqref{equ:pcmm}, the point-correlation measurement model can be expressed as
\begin{equation}
\mathbf{z}_{ijk} = \mathbf{y}_{ik} - \mathbf{y}_{jk}=  \mathbf{s}_{RGB-D} ( \mathbf{T}_k \cdot (\overline{\mathbf{p}}_{i} - \overline{\mathbf{p}}_{j} ))  + \mathbf{n}_{ijk},
\label{equ:meas_con_imp}
\end{equation}
where $\mathbf{z}_{ijk}$ is the relative position between $\mathbf{p}_{i}$ and $\mathbf{p}_{j}$ for RGB-D sensors.
If we assume that $\mathbf{y}_{ik}$ and $\mathbf{y}_{jk}$ are independent, the covariance of $\mathbf{z}_{ijk}$ between the two points can be computed as
\begin{equation}
\mathbf{C}_{ijk} =  \mathbf{C}_{ik}+\mathbf{C}_{jk}.
\end{equation}

\subsubsection{Determination of Static Points by the Front End}
The static points are determined by the front end to eliminate the influence of fast-moving objects on its motion estimation.
Because incorrect correspondences affect the optimization of the point correlations, and the pose estimation step can remove false correspondences, the static point determination module is added after initial pose estimation.
As introduced in Section \ref{sec:segmentation_init}, the 3D graph $\mathcal{G} = \{\mathbf{l}_{ij}\}$ is constructed using Delaunay triangulation based on the tracked map points of the previous frame. Only the measurements of the two frames are used to optimize the graph. Then, the squared Mahalanobis length of the error is computed for each measurement of each edge $\mathbf{l}_{ij} \in \mathcal{G}$.
If the error in a measurement is larger than the given threshold, this measurement is removed.
If all measurements of an edge are removed, the edge is determined to be an inconsistent edge and is removed from $\mathcal{G}$.
After removing all inconsistent edges, the remaining graph $\mathcal{G}_{new}$ is separated into multiple connected components.
Afterward, all connected components are found by checking the connectedness of $\mathcal{G}_{new}$, and the result is denoted by $\{\mathcal{C}_i\}$.
Finally, the points of the largest connected component of $\{\mathcal{C}_i\}$ are determined to be reliable static points, and the marks of every map point in $\{ \mathcal{G}_{new} \setminus \mathcal{C}_{max}\}$ are removed.

\subsubsection{Static Point Determination in the Back End}

\label{Sec:FGS}

\begin{algorithm}
	\caption{Back-End Segmentation }
	\begin{algorithmic}[1] 
		\REQUIRE   - Static local map-point set $\mathcal{P}$ 
		\\	- Maximum number of iterations $n$ allowed in the algorithm
		\\  - Threshold value $t$
		\ENSURE - Graph $\mathcal{G}_{output}$
		\STATE Triangulate map points $\mathcal{P}$ to obtain the 3D edge set $\mathcal{G} = \{\mathbf{l}_{ij}\}$
		\FOR{ $\mathbf{l}_{ij} \in \mathcal{G} $}
		\STATE $\mathbf{y}_{i} \gets$ the observation set \{$\mathbf{y}_{ik}$\} of $\mathbf{p}_i$
		\STATE $\mathbf{y}_{j} \gets$ the observation set \{$\mathbf{y}_{jk}$\}  of $\mathbf{p}_j$
		\IF {$\mathbf{y}_{i}$ and $\mathbf{y}_{j}$ have the same observation in frame $k$}
		\STATE Compute the observation $\mathbf{z}_{ijk}$ of $\mathbf{l}_{ij}$ and  $\mathbf{C}_{ijk}$
		\STATE Add $\mathbf{z}_{ijk}$ to $\mathbf{z}$ 
		\ENDIF
		\ENDFOR 
		
		\STATE $\mathbf{z}_{inlier} \gets \mathbf{z}$
		\STATE $iterations \gets 0$
		\FOR  {$iterations < n$}

		\STATE Optimize the objective function in Eq. (\ref{equ:geo_prior}) with $\mathbf{z}_{inlier}$
		\FOR{ $\mathbf{z}_{ijk} \in \mathbf{z}$}
		\IF{ $\mathbf{e}_{z,ijk}(\mathbf{l}_{ijk})^T \mathbf{C}^{-1}_{ijk} \mathbf{e}_{z,ijk}(\mathbf{l}_{ijk}) >  t$  }
		\STATE Remove $\mathbf{z}_{ijk}$ from $\mathbf{z}_{inlier}$ 
		\ELSE
		\STATE Add $\mathbf{z}_{ijk}$ to $\mathbf{z}_{inlier}$ 
		\ENDIF
		\ENDFOR 
		\STATE $iterations \gets iterations+1 $
		\ENDFOR 
		\STATE Compute a new graph $\mathcal{G}_{new}$ based on the remaining $\mathbf{z}_{inlier}$
		\STATE Determine whether the graph $\mathcal{G}_{new}$ is separated into several connected components \{$\mathcal{C}_i$\} using DFS 
		\STATE Compute the volume of each connected component $\mathcal{C}_i$ and determine the largest one $\mathcal{C}_{max}$
		\STATE Erase the static marks of each map point in $ \{ \mathcal{G}_{new} \setminus \mathcal{C}_{max}\}$
		\STATE $\mathcal{G}_{output} \gets \mathcal{C}_{max}$
	\end{algorithmic}
	\label{agl:back-end}
\end{algorithm}

This module verifies the marked points after new keyframes have been added. 
The calculation is shown in Algorithm \ref{agl:back-end}.
In contrast to the determination module in the front end, this module attempts to use all of the information in the sliding window to determine the static points.

This module searches for the keyframes that have covisibility with the currently added keyframe. Then, a 3D graph structure is created using the marked points of these obtained keyframes. 
The available measurements of all edges are used to build the Hessian matrix of the objective function in Eq. (\ref{equ:geo_prior}). Afterwards, the outliers are removed in the iterations.
After optimization and the removal of inconsistent edges from the graph, the remaining graph is divided into multiple connected components if there are moving objects in the FOV. 
Finally, the points of the component with the largest volume are retained as marked points, but the points in the smaller connected components are changed to unmarked points.

\subsection{Map-Point Matching}

\label{sec:matching}

\begin{figure}[t]
	\centering
	\includegraphics[width=0.4 \textwidth]  {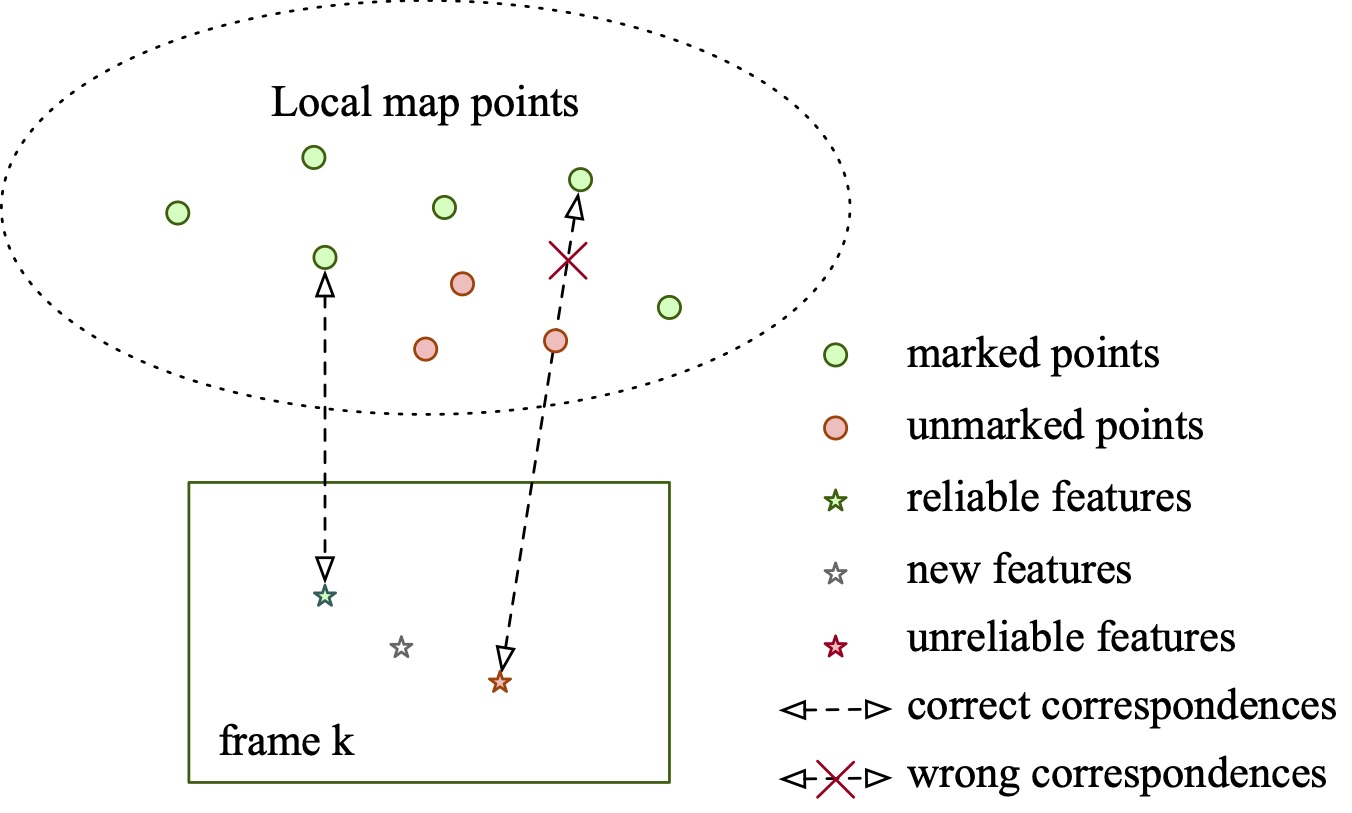}
	\caption{Example of map-point matching. If the unmarked points are removed immediately after the static point determination module, some unreliable features on the image will incorrectly match the marked points.}
	\label{fig:matching}
\end{figure}

Map-point matching is used to obtain the matching between map points and features.
As shown in Figure \ref{fig:matching}, the matching of marked points can determine the features that are reliable in the static scene in the FOV of the frame. 
In contrast, the matching of unmarked points can include features that are not reliable and should not be used for motion estimation.
In addition, the determination of unreliable features reduces the risk of failure when identifying static points and the probability of incorrect matching.

\subsection{Map Management}
Map management is divided into two parts: point management and keyframe management.

\subsubsection{Point Management}
Point management consists of three tasks: creation, updating, and culling.

\textbf{Creation:}
There are two methods for creating map points.
In the first method, map points are created by triangulation from two keyframes that have covisibility. 
In the second method, map points are immediately created from a new feature that has a depth measurement.
For the first method, it is difficult to create a point on a moving object because of the huge triangulation error.
For the second method, a point on a moving object will be created because the corresponding new feature cannot be recognized as a feature on the moving object from a single image.
As described in Section \ref{sec:matching}, dynamic points are prevented from being created by the second method because the features are matched with the unmarked points.

\textbf{Update:} After creation, updating the positions of the marked points improves the accuracy of motion estimation by the front end and can avoid mismatches. Therefore, the marked and unmarked points are separately updated using new information.
The marked points are updated by bundle adjustment.
The unmarked points are updated by static point determination using the newly available measurements from the new frame.

\textbf{Culling:}
In addition to creation, the map points must meet a criterion to ensure that no point is incorrectly triangulated.
Each type of map point has its own criterion.
Marked points must have a certain number of successful observations,  which ensures that they are trackable. Otherwise, they should be removed.
The unmarked points are removed when they are not observed in the FOV of the sliding window.

\subsubsection{Keyframe Management}

Keyframe culling is used to maintain a compact reconstruction.
The policy is that the keyframes in which most points have been observed in other keyframes are discarded.
Keyframe culling confers two benefits.
\textbf{The preservation of valid data:} Frames without new static information are often mistakenly identified as keyframes owing to the presence of moving objects. Therefore, these incorrect keyframes are more likely to be discarded after the static points have been correctly determined.
\textbf{A longer sliding-window time span:} Discarding redundant keyframes lengthens the time spans of the sliding windows without increasing the number of keyframes. A longer time span increases the ability of the static point determination module to eliminate the influence of slow-moving objects.

\section{Experiments}
\label{sec:expreiments}

\begin{table*}[t]
	\centering
	\caption{Comparison of the rotational root mean-squared error (RMSE) of the relative pose error (RPE) on the TUM benchmark. The best results are shown in bold. Not all papers provide results for all sequences. We report the improvement with respect to the original SLAM system (ORB-SLAM2) without static point determination (SPD).
	}
	\begin{tabular}{cc|cccc|ccc}
		\toprule
		\multicolumn{2}{c|}{\multirow{2}{*}{Sequences}}    & \multicolumn{7}{c}{Rot. RMSE of trajectory alignment [$^\circ/s$]}    \\
		\multicolumn{2}{c|}{}    & BAMVO  & StaticFusion  & SPWSLAM & DVO     & \begin{tabular}{@{}c@{}}
     ORB-SLAM2   \\
     w/o SPD
    \end{tabular}%
		&  \begin{tabular}{@{}c@{}}
     Our \\
     w/ SPD
    \end{tabular}%
    & \begin{tabular}{@{}c@{}}
     Improvement \\
     w/ SPD
    \end{tabular}%
    \\
		\midrule
		\multirow{5}{*}{slightly dynamic}  & fr2/desk-person  & 1.2159 & - & \textbf{0.8213}  & 1.5368 & 1.3717  & 1.3951 & -1.70\% \\
		& fr3/sitting-static      & 0.6997 &0.43 & 0.7228 & 0.6084 &   \textbf{0.3630} &   0.3786  & -4.30\%  \\
		& fr3/sitting-xyz         & 1.3885 &0.92 & 0.8466 & 1.4980 &   0.5817 &  \textbf{0.5792} & 0.43\% \\
		& fr3/sitting-rpy         & 5.9834 &- & 5.6258 & 6.0164 &   0.9361  &    \textbf{0.9047} & 3.35\%  \\
		& fr3/sitting-halfsphere  & 2.8804 &2.11 & 1.8836 & 4.6490 &   0.9101   &    \textbf{0.8699}  & 4.41\% \\
		\midrule
		\multirow{4}{*}{highly dynamic} & fr3/walking-static           & 2.0833 &0.38 & 0.8085 & 6.3502 &     10.5764    & \textbf{0.3293} & 96.89\% \\
		& fr3/walking-xyz           & 4.3911 &2.66 & \textbf{1.6442}   & 7.6669 &      19.7299   & 2.7413 & 86.11\% \\
		& fr3/walking-rpy           & 6.3398 &- & {5.6902}  & 7.0662 &        22.2934 &   \textbf{4.6327}  & 79.22\%  \\
		& fr3/walking-halfsphere    & 4.2863 &5.04 & 2.4048 & 5.2179 &    24.6634   &   \textbf{0.9854} & 96.00\%  \\
		\bottomrule
	\end{tabular}
\label{tab:rpe_rota}
\end{table*}

\begin{table*}[t]
	\centering
	\caption{Comparison of the translational RMSE of the RPE on the TUM benchmark. The best results are shown in bold. Not all papers provide results for all sequences. We report the improvement with respect to the original SLAM system (ORB-SLAM2) without static point determination (SPD).
	}
	\resizebox{18cm}{!}
	{
	\begin{tabular}{cc|ccccc|ccc}
		\toprule
		\multicolumn{2}{c|}{\multirow{2}{*}{Sequences}}         & \multicolumn{8}{c}{ Trans. RMSE of trajectory alignment [m/s]}    \\
		\multicolumn{2}{c|}{}        & BAMVO  & StaticFusion & FlowFusion & SPWSLAM & DVO     & \begin{tabular}{@{}c@{}}
     ORB-SLAM2   \\
     w/o SPD
    \end{tabular}%
		&  \begin{tabular}{@{}c@{}}
     Our \\
     w/ SPD
    \end{tabular}%
    & \begin{tabular}{@{}c@{}}
     Improvement \\
     w/ SPD
    \end{tabular}%
    \\
		\midrule
		\multirow{5}{*}{slightly dynamic}  & fr2/desk-person            & 0.0352 &- & - & \textbf{0.0173} & 0.0354 &         0.0377    & 0.0362 & 3.98 \% \\
		& fr3/sitting-static          & 0.0248 &\textbf{0.011} & - & 0.0231 & 0.0157 &           0.0122  &   0.0138  & -13.11\%  \\
		& fr3/sitting-xyz              & 0.0482 &0.028 & - & 0.0219 & 0.0453 &        0.0137    &  \textbf{0.0134} & 2.19\% \\
		& fr3/sitting-rpy             & 0.1872 &- & - & 0.0843  & 0.1735 &        0.0380    &    \textbf{0.0320}  & 15.79\%  \\
		& fr3/sitting-halfsphere         & 0.0589 &0.030 & - & 0.0389 & 0.1005 &      0.0365    &    \textbf{0.0354} & 3.01\%  \\
		\midrule
		\multirow{4}{*}{highly dynamic} & fr3/walking-static         & 0.1339 &\textbf{0.013} & 0.030 & 0.0327  & 0.3818 &     0.5826     & 0.0141 & 97.58\% \\
		& fr3/walking-xyz                 & 0.2326 &0.121 & 0.21 & \textbf{0.0651}   & 0.4360 &      1.0484     & 0.1266  & 87.92\% \\  
		& fr3/walking-rpy             & 0.3584 & - & - & \textbf{0.2252} & 0.4038 &          1.1843    &   0.2299  & 80.59\%  \\
		& fr3/walking-halfsphere          & 0.1738 & 0.207 & - & 0.0527 & 0.2628 &    1.0790     &   \textbf{0.0517}  & 95.21\% \\
		\bottomrule
	\end{tabular}
	}
	\label{tab:rpe_trans}
\end{table*}

\begin{table*}[t]
	\centering
	\caption{Comparison of the absolute trajectory error (ATE) on the TUM benchmark. The best results are shown in bold. Not all papers provide results for all sequences. We report the improvement with respect to the original SLAM system (ORB-SLAM2) without static point determination (SPD).
	}
	\label{tab:ate}
	\resizebox{18cm}{!}
	{
	\begin{tabular}{cc|ccccc|ccc}
		\toprule
		\multicolumn{2}{c|}{\multirow{2}{*}{Sequences}}         & \multicolumn{8}{c}{ Trans. RMSE of  trajectory alignment [m]}    \\
		\multicolumn{2}{c|}{}       &  DVO SLAM   &   \begin{tabular}{@{}c@{}}
     DVO SLAM   \\
     Motion Removal
    \end{tabular}%
    & StaticFusion & FlowFusion & SPWSLAM &  \begin{tabular}{@{}c@{}}
     ORB-SLAM2   \\
     w/o SPD
    \end{tabular}%
		&  \begin{tabular}{@{}c@{}}
     Our \\
     w/ SPD
    \end{tabular}%
    & \begin{tabular}{@{}c@{}}
     Improvement \\
     w/ SPD
    \end{tabular}%
     \\
		\midrule
		\multirow{5}{*}{slightly dynamic}  & fr2/desk-person      &0.1037    &0.0596   &- & - & 0.0484  &         \textbf{0.0064}  & 0.0075 & -17.18\% \\
		& fr3/sitting-static     &0.0119  &-    &0.013  & - & - & \textbf{0.0077} &   0.0096 & -24.68\%  \\
		& fr3/sitting-xyz        &0.2420    &0.0482   &0.040 & - & 0.0397   &        0.0094   &  \textbf{0.0091} & 3.19\% \\
		& fr3/sitting-rpy       &0.1756  & -   & -    &- & -  &        0.0250  &    \textbf{0.0225}  & 10.0\% \\
		& fr3/sitting-halfsphere  &0.2198   & 0.1252        &0.040 & - & 0.0432  &      0.0250 &    \textbf{0.0235} & 6.00\%  \\
		\midrule
		\multirow{4}{*}{highly dynamic} & fr3/walking-static   &0.7515   & 0.0656 &0.014 & 0.028 & 0.0261 &     0.4080    & \textbf{0.0108} & 97.35\% \\
		& fr3/walking-xyz      &1.3830     & 0.0932   &0.127  & 0.12 & \textbf{0.0601}  &      0.7215 & 0.0874 & 87.88\% \\ 
		& fr3/walking-rpy     &1.2922     & 0.1333  &- & - & 0.1791 &  0.8054 &  \textbf{0.1608}  & 80.03\%  \\
		& fr3/walking-halfsphere  &1.0136 & 0.470   &0.391  & - & 0.0489 &    0.7225  &   \textbf{0.0354}  & 95.10\% \\
		\bottomrule
	\end{tabular}
	}
\end{table*}

\begin{table*}[t]
	\centering
	\caption{Comparison of the ATE on the TUM benchmark. The best results are shown in bold. Not all papers provide results for all sequences. $^*$ indicates the methods that segment the scene into different objects using both motion and semantic cues (deep learning).
	}
	\label{tab:ate_learning}
	{
	\begin{tabular}{cc|ccccc|c}
		\toprule
		\multicolumn{2}{c|}{\multirow{2}{*}{Sequences}}         & \multicolumn{6}{c}{ Trans. RMSE of  trajectory alignment [m]}    \\
		\multicolumn{2}{c|}{}      & Co-Fusion$^*$ & MaskFusion & MID-Fusion & EM-Fusion  & DynaSLAM$^*$  & Our   \\
		\midrule
		\multirow{5}{*}{slightly dynamic}  & fr2/desk-person &-&-&- & -& -   & \textbf{ 0.0075} \\
		& fr3/sitting-static  &0.011&0.021& 0.010 & \textbf{0.09} & - &   \textbf{0.0096}   \\
		& fr3/sitting-xyz     &0.027&0.031& 0.062 & 0.37   &  0.015     &  \textbf{0.0091} \\
		& fr3/sitting-rpy     &-&-&-   & - &-  &  \textbf{0.0225}   \\
		& fr3/sitting-halfsphere &0.036& 0.052 &0.031 & 0.032 &      \textbf{0.017}  &    {0.0235}   \\
		\midrule
		\multirow{4}{*}{highly dynamic} & fr3/walking-static   &0.551&0.035& 0.023  & 0.014  &     \textbf{0.006}   & {0.0108} \\
		& fr3/walking-xyz      &0.696 &0.104 &0.068  & 0.066  &     \textbf{0.015}  & 0.0874 \\ 
		& fr3/walking-rpy     &-&-&-   &-   &   \textbf{ 0.035} &  {0.1608}    \\
		& fr3/walking-halfsphere &0.803 &0.106 &0.038 & 0.051  &  \textbf{ 0.025 } &   {0.0354}   \\
		\bottomrule
	\end{tabular}
	}
\end{table*}

 \begin{figure*}[t]
 	\centering
 	\includegraphics[scale=0.5]  {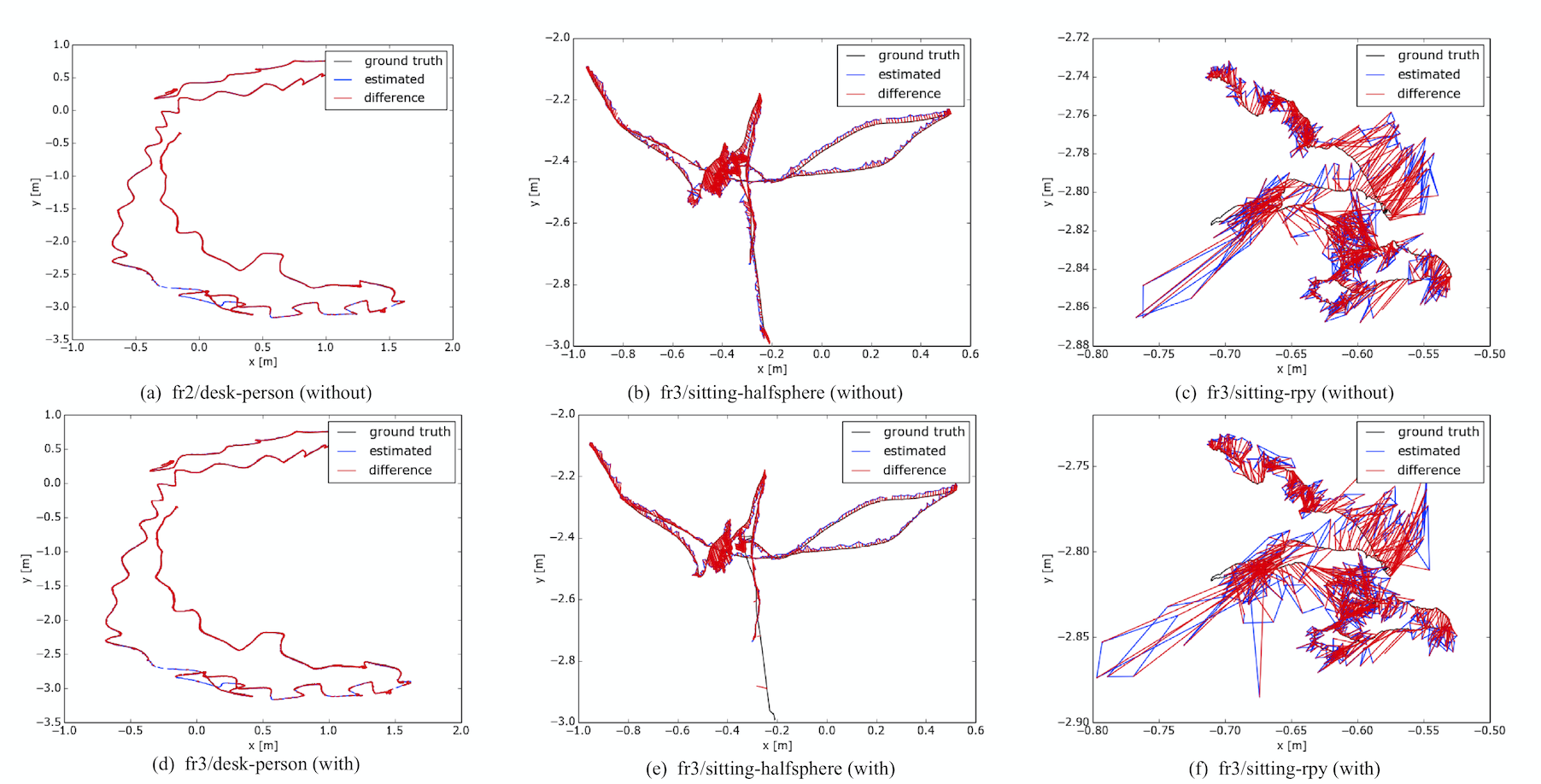}
 	\caption{Comparison of example estimated trajectories. (a), (b), (c) Trajectories estimated with ORB-SLAM2, which the proposed method is built on and is not designed for dynamic environments. (d), (e), (f) Trajectories estimated with the proposed method.}
 	\label{fig:compare1}
 \end{figure*}

 \begin{figure*}[t]
 	\centering
 	\includegraphics[scale=0.5]  {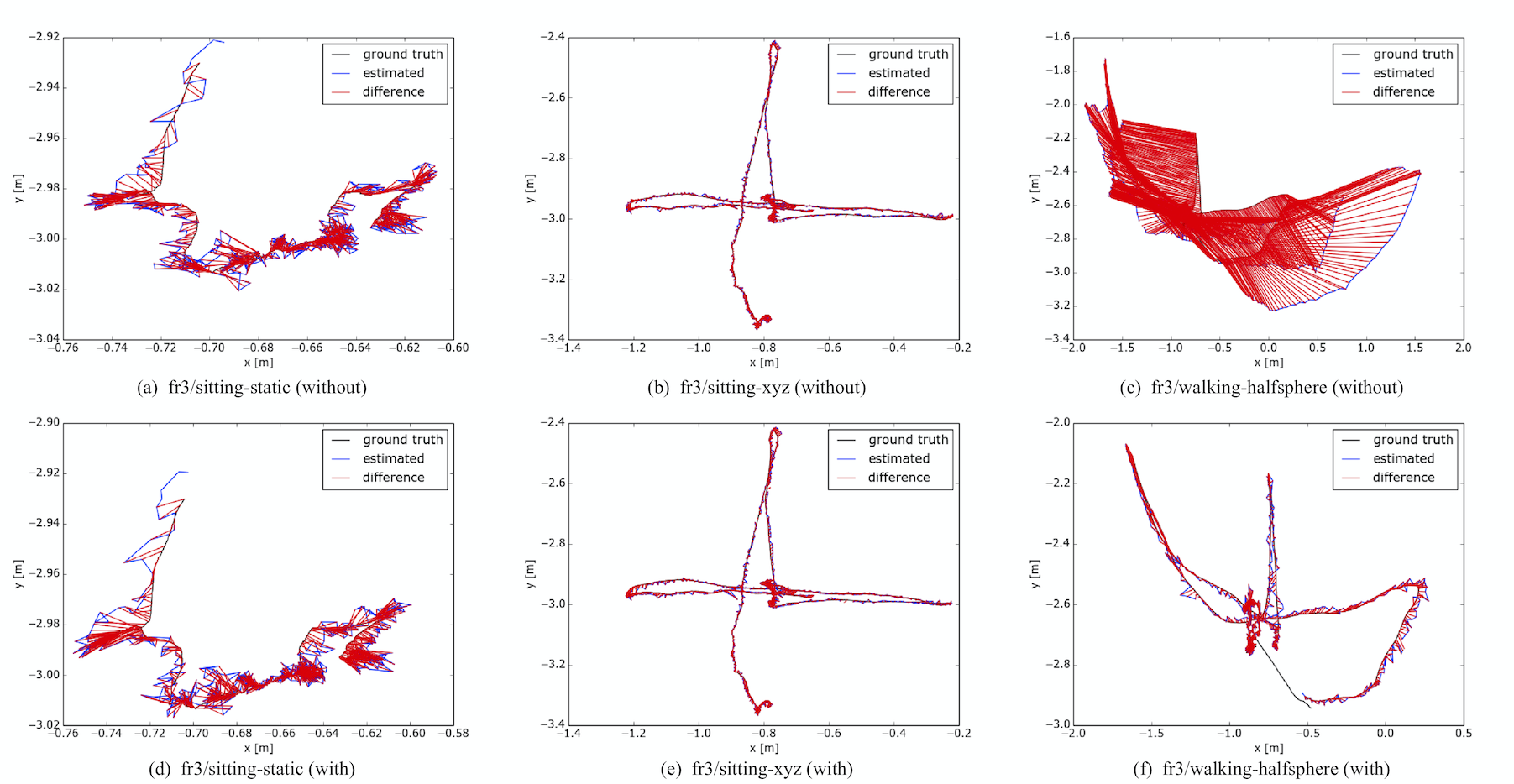}
 	\caption{Comparison of example estimated trajectories. (a), (b), (c) Trajectories estimated with ORB-SLAM2, which the proposed method is built on and is not designed for dynamic environments. (d), (e), (f) Trajectories estimated with the proposed method.}
 	\label{fig:compare2}
 \end{figure*}

\begin{figure*}[t]
	\centering
	\includegraphics[scale=0.5]  {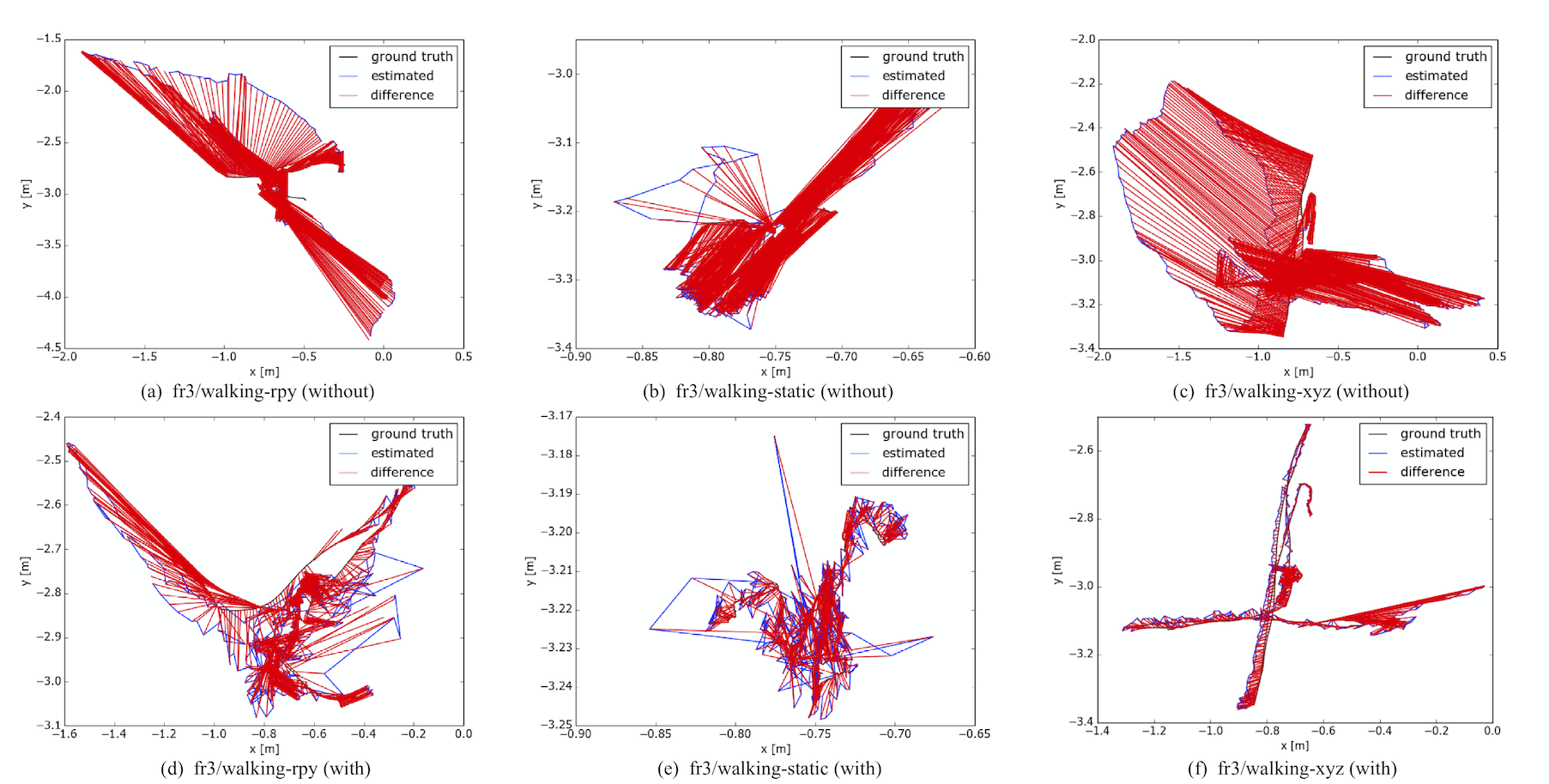}
	\caption{Comparison of example estimated trajectories. (a), (b), and (c) Trajectories estimated using ORB-SLAM2, which is not designed for dynamic environments. (d), (e), and (f) Trajectories estimated using the proposed method.}
	\label{fig:compare3}
\end{figure*}

 \begin{figure*}[t]
 	\centering
 	\includegraphics[scale=0.5]  {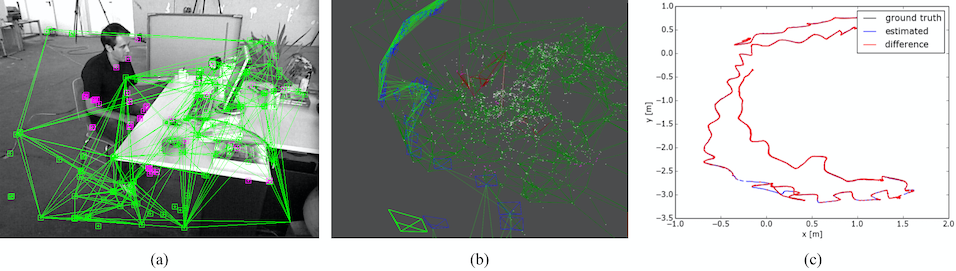}
 	\caption{Example taken from the fr2/desk-with-person sequence. (a) Result showing the determination of static points by the front end. (b) Result of 3D edge culling during the determination of feature static points by the back end. (c) Estimated trajectory compared to the ground truth. In (a), the green line is the stable correlation between two points. If the points are determined on the moving object, the tracked feature result and 3D visualization are marked in pink. In (b), the red line is the edge between dynamic points in three dimensions, and the green line is the edge between static points.}
 	\label{fig:f2_exp}
 \end{figure*}

 \begin{figure*}[t]
 	\centering
 	\includegraphics[scale=0.5]  {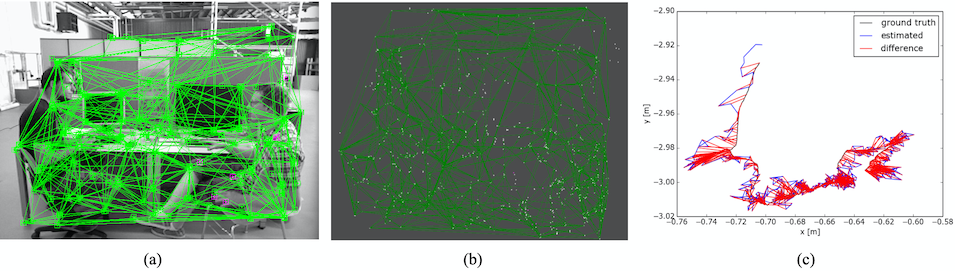}
 	\caption{Example taken from the fr3/sitting-static sequence. (a) Result showing the determination of static points by the front end. (b) Result of 3D edge culling during the determination of feature static points by the back end. (c) Estimated trajectory compared to the ground truth.   }
 	\label{fig:sitting_static_exp}
 \end{figure*}

\begin{figure*}[t]
	\centering
	\includegraphics[scale=0.5]  {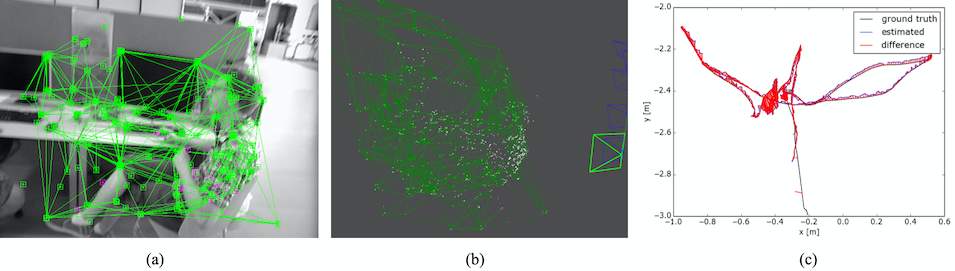}
	\caption{Example taken from the fr3/sitting-halfsphere sequence. (a) Result showing the determination of static points by the front end. (b) Result of 3D edge culling during the determination of feature static points by the back end. (c) Estimated trajectory compared with the ground truth.  }
	\label{fig:sitting_half_exp}
\end{figure*}

 \begin{figure*}[t]
 	\centering
 	\includegraphics[scale=0.5]  {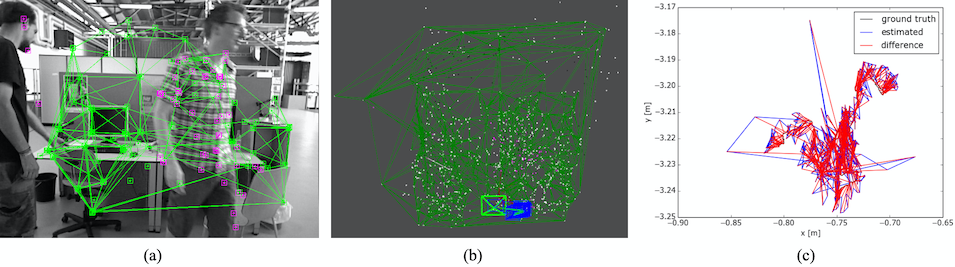}
 	\caption{Example taken from the fr3/walking-static sequence. (a) Result showing the determination of static point by the front end. (b) Result of 3D edge culling during the determination of feature static points by the back end. (c) Estimated trajectory compared to the ground truth.}
 	\label{fig:walking_static_exp}
 \end{figure*}

\begin{figure*}[t]
	\centering
	\includegraphics[scale=0.5]  {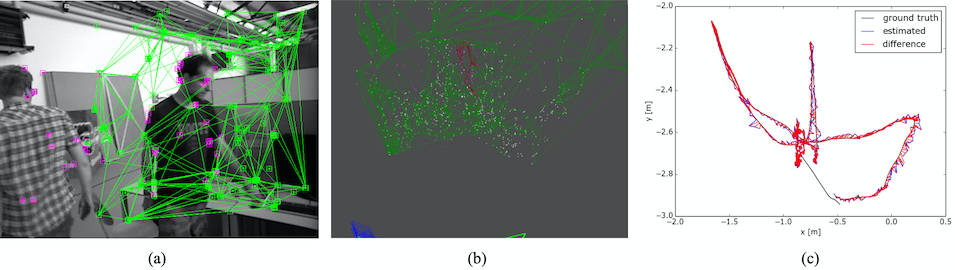}
	\caption{Example taken from the fr3/waling-halfsphere sequence. (a) Result showing the determination of static points by the front end. (b) Result of 3D edge culling during the determination of feature static points by the back end. (c) Estimated trajectory compared with the ground truth.}
	\label{fig:walking_half_exp}
\end{figure*}

In this section, we present an experimental evaluation with indoor sequences from the Technical University of Munich (TUM) RGB-D benchmark \cite{sturm2012benchmark} to evaluate the accuracy of DSLAM as well as three sequences from especially challenging environments to evaluate the robustness of DSLAM.
We also extracted the statistics of the time spent in each step of the static point determination module to evaluate the efficiency of the proposed segmentation method.

The TUM benchmark includes ground-truth trajectories obtained from a high-accuracy motion capture system and contains both the static and dynamic scenarios in indoor environments. 
We divided the environments of the TUM benchmark into three categories: static, slightly dynamic, and highly dynamic. When there are no moving objects in the scene, we call it a \textsl{static environment}. 
If only a small part of the FOV is covered by moving objects, e.g., someone in the office makes a gesture, it is defined as a \textsl{slightly dynamic environment}. 
If the majority of the FOV is occupied by moving objects, we call it a \textsl{highly dynamic environment}.
The performance of most state-of-the-art RGB-D methods are evaluated on the TUM RGB-D dataset, and these methods have achieved good results. However, the sequences containing moving objects are not often used for evaluation. Therefore, the slightly and highly dynamic types of sequences were used to evaluate the performance of the proposed method in dynamic environments.

All experiments were performed on a desktop computer equipped with an Intel Core i5-3470 (3.2 GHz) CPU and 8 GB of RAM. 
For comparison, we also obtained  results for the following state-of-the-art methods: dense VO (DVO) \cite{kerl2013robust}, ORB-SLAM2 \cite{sun2017improving}, model-based dense VO (BAMVO) \cite{kim2016effective},and DVO SLAM \cite{kerl2013dense}. In addition, the reported results of  FlowFusion\cite{FlowFusion2020Zhang}, Motion Removal DVO SLAM \cite{sun2017improving}, StaticFusion\cite{scona2018staticfusion}, RGBD SLAM with static-point weighting (SPWSLAM) \cite{li2017rgb}, Co-Fusion \cite{runz2017co}, MaskFusion \cite{runz2018maskfusion}, MID-Fusion \cite{xu2019mid}, EM-Fusion\cite{strecke2019fusion}, and DynaSLAM \cite{bescos2018dynslam} are included in the comparison. 
The results for the improvements to the original SLAM system without the proposed segmentation are also reported to determine whether the improvements in the results are due to the segmentation method or the SLAM core systems.
DVO, DVO SLAM, and ORB-SLAM2 represent the most advanced methods based on the static world assumption. Motion Removal DVO SLAM, BAMVO, FlowFusion, StaticFusion, and SPWSLAM are recent methods that consider the influence of moving objects. Finally, Co-Fusion, MaskFusion, MID-Fusion, EM-Fusion, and DynaSLAM are methods that leverage learning techniques.

\subsection{Comparison of the Accuracy}

In the TUM benchmark, the slightly dynamic environment sequences are the \textsl{sitting} and \textsl{desk-person} sequences, and the highly dynamic environment sequences are the \textsl{walking} sequences. The \textsl{walking} sequences are challenging because the moving objects cover a large part of the FOV. In the slightly dynamic environment, there is a person sitting in front of a desk and moving their arms, sometimes in an organized office. In the highly dynamic environment, two people are walking around a desk. In the sequences for both types of environments, there are four types of camera motion, which are indicated in the sequence names. Here, \textsl{halfsphere} indicates that the camera follows the trajectory of a 1-m diameter half sphere, \textsl{xyz} indicates that the camera almost moves along the $x$, $y$, and $z$ axes, \textsl{rpy} indicates that the camera rotates in the roll, pitch, and yaw directions, and \textsl{static} indicates that the camera only moves around a position in the environment.

The translational RMSE of the RPE in meters per second and the rotational RMSE of the RPE in degrees per second were calculated for the evaluation. The RMSE of the RPE is much more easily influenced by large occasional errors in the estimate; thus, it is more suitable for an evaluation in dynamic environments. The results for the RPE are listed in Tables \ref{tab:rpe_rota} and \ref{tab:rpe_trans}. 
We also evaluated the full trajectory performance of the proposed method using the translational RMSE of the ATE, as listed in Table \ref{tab:ate}. 
Moreover, a comparison with methods that use learning techniques is shown in Table \ref{tab:ate_learning}.
The estimated trajectories were compared with the ground truth, and some results obtained by the proposed method are shown in Figures \ref{fig:sitting_half_exp} and \ref{fig:walking_half_exp}. It can be seen that the proposed method is able to process all sequences well, including those in both slightly and highly dynamic environments.

\textbf{Slightly dynamic environments:} Compared to the methods with the static world assumption, DSLAM is slightly more accurate.
In \textsl{fr3/desk-person}, the methods using the static world assumption provide slightly better results because most of the influence of moving objects can be eliminated.
However, as the proportion of moving objects occupying the FOV increases, the robust estimation methods are not able to discard all dynamic points.
Therefore, DSLAM provides better results in the \textsl{sitting} sequences.

In addition, DSLAM is able to outperform the methods that consider moving objects.
The performances of methods that consider moving objects are even worse than those obtained using ORB-SLAM2.
The reason is that a part of the information of the static scene is mistakenly removed by aggressive thresholds or image segmentation methods in these methods. Therefore, there is less available static information for motion estimation, decreasing accuracy. The poor performance in slightly dynamic environments limits their application in practice. DSLAM, which makes use of static points, provides the best results in most sequences.

Moreover, when compared with the learning-based methods, the proposed method performs better, as shown in Table \ref{tab:ate_learning}. In the slightly dynamic environment with people sitting in front of the desk, most of the body is static, and only their hands move. Because the entire image region of static movable objects is detected and ignored, most static information in the person is discarded. As a result, less static information is used by learning-based methods to estimate the camera pose.

\textbf{Highly dynamic environments:} Because the static world assumption is not true, DSLAM outperforms the methods that assume a static world.
In these sequences, most of the features on the moving objects are incorrectly tracked as inliers by these methods, which are not designed for dynamic environments.
Therefore, both DVO and ORB-SLAM2 yield unacceptable errors. 
The results for ORB-SLAM2 show that loop-closure detection does not correctly reduce drift because the moving objects dramatically influence the loop closing module.

When compared with the methods that consider moving objects, DSLAM also provides better results for most sequences. 
DSLAM performs worse than SPWSLAM in only one case, namely the \textsl{walking-xyz} sequence.
In the beginning part of the \textsl{walking-xyz} sequence, the people walk away from the camera.
Therefore, the position of the person in the RGB image only slightly changes, while the depth value of the person in the depth images substantially changes.
Meanwhile, because depth measurements are not very accurate, the segmentation relies more on the information of the RGB image in our implementation.
Therefore, it is difficult to separate dynamic points when segmentation is performed with an insufficient amount of information from RGB images.
This problem can also be seen in the results shown in Figure \ref{fig:compare3}, where a significant error is introduced in the beginning part of the sequence.
This problem can be solved by setting the threshold value aggressively. However, more aggressive parameters will lead to more false negatives, which reduces the accuracy of the results obtained on other sequences.

In this highly dynamic environment, the features on the bodies of people are all moving dynamically and should be excluded. Because the entire image region of the person is masked using learning techniques, the learning-based methods only utilizing static information should perform better. However, the proposed method performs better than most learning-based methods. Only DynaSLAM can perform better than DSLAM. Therefore, the motion estimation core system and geometry cues are also crucial for increasing accuracy.

In the comparison of several methods that use learning techniques (Co-Fusion, MaskFusion, MID-Fusion, EM-Fusion, and DynaSLAM), the results differ even when using the same neural network, as shown in Table \ref{tab:ate_learning}. DynaSLAM, the learning-based method that combines geometry cues, performs the best at present. The reason is that though only people appear in the benchmark environment as moving objects, the learning-based techniques still cannot obtain 100\% accuracy because of the nature of the algorithms. 
In addition, there are objects that cannot be detected by learning-based techniques because they are not a priori dynamic, but they are movable. Therefore, the methods that only depend on learning techniques do not offer excellent performance, even in environments that only contain pretrained objects. Furthermore, in the presence of unknown objects that were not present in the training set, these learning-based methods may fail. This poor generalization limits their use in unknown environments. Moreover, most learning methods require pre-training in advance for better performance, which makes them even more inconvenient to use in practice.

\subsection{Comparison of the Robustness}

\begin{figure*}[t]
	\centering
	\includegraphics[scale=1.0]  {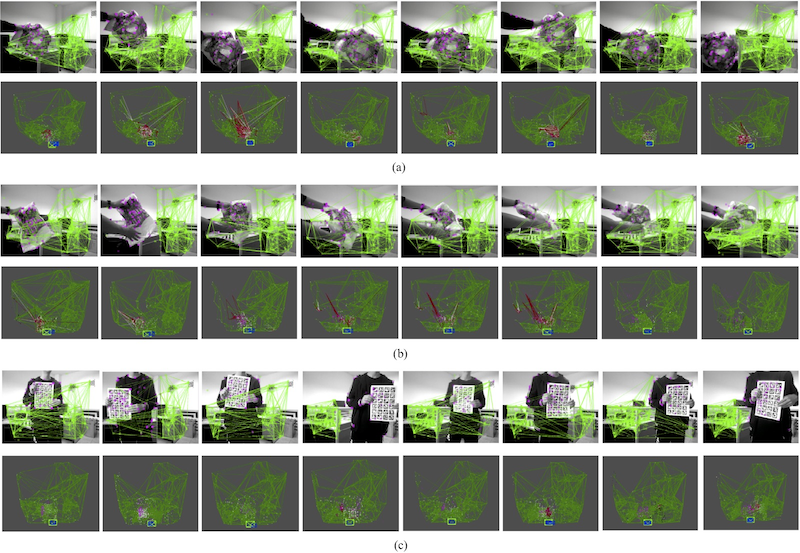}
	\caption{Tracking and static point determination results for each of the three challenging environments: (a) environment with an object moving at a slow speed, (b) environment with an object moving with unstructured motion, and (c) environment with a person walking and covering most of the FOV.}
	\label{fig:comparsion_ownofficeseq}
\end{figure*}
\begin{figure}[t]
	\centering
	\includegraphics[scale=0.15]  {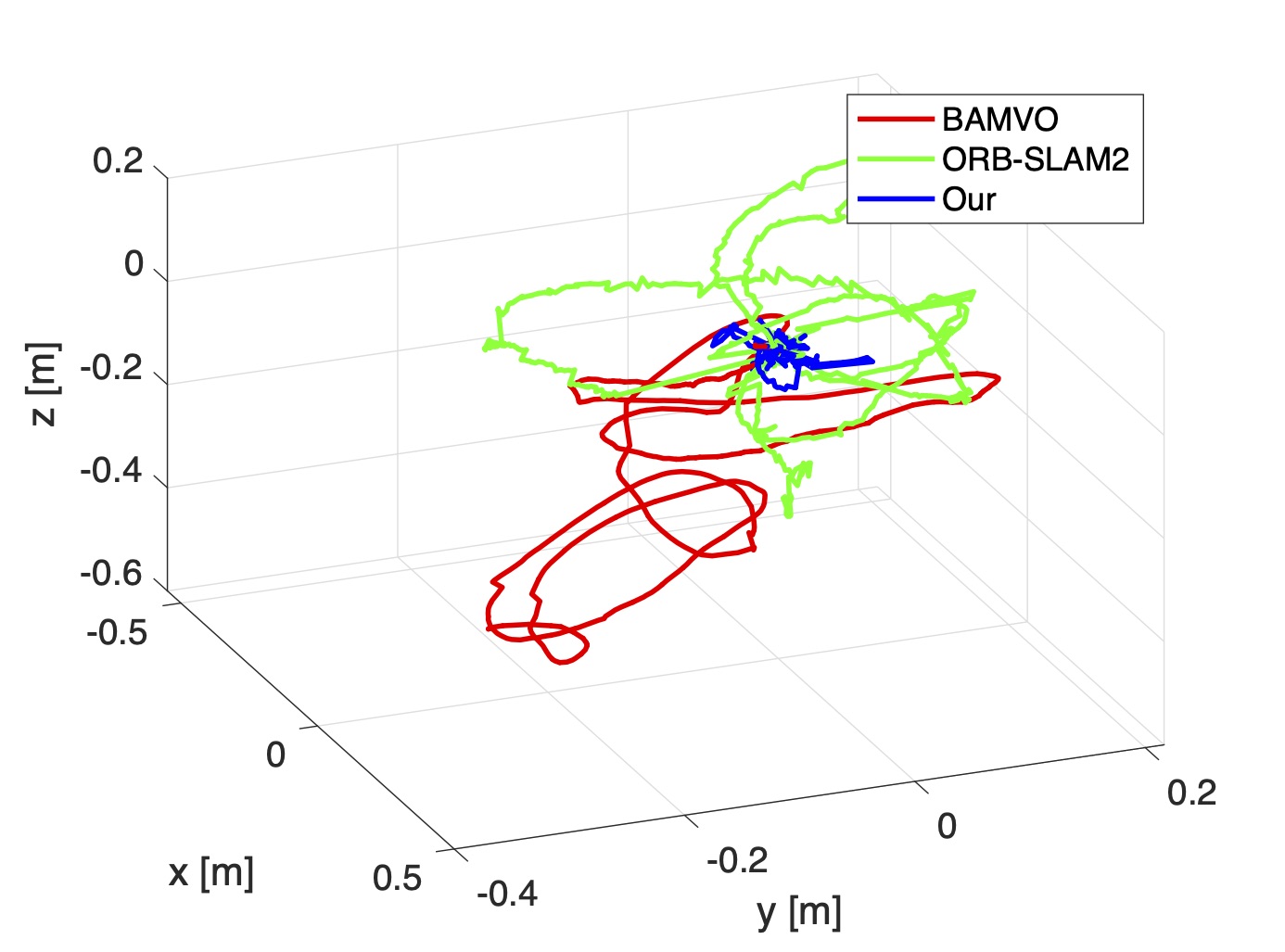}
	\caption{XYZ trajectory results for the environment with an object moving at a slow speed.}
	\label{fig:comparsion_ownoffice1}
\end{figure}
\begin{figure}[t]
	\centering
	\includegraphics[scale=0.15]  {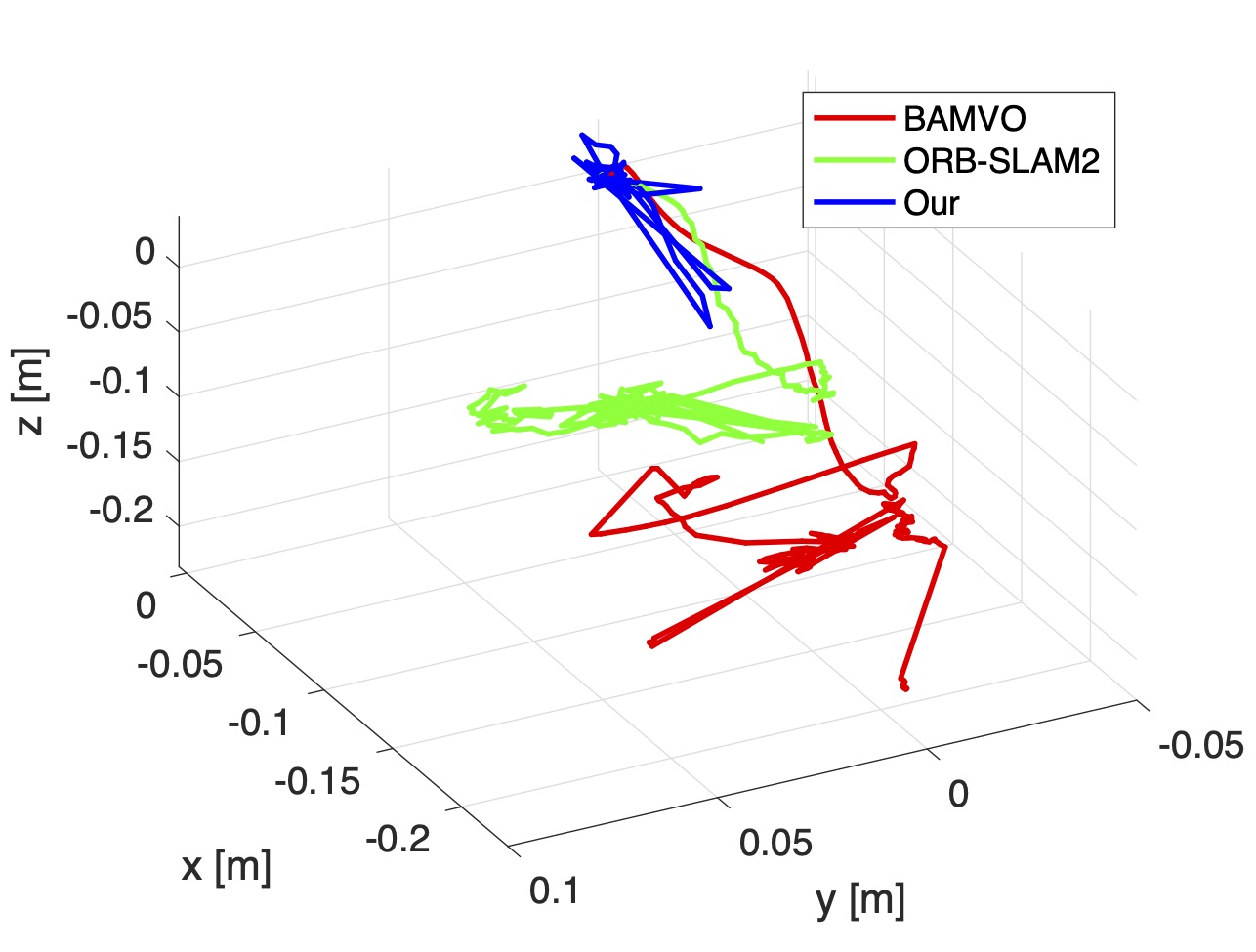}
	\caption{XYZ trajectory results for the environment with an object moving with unstructured motion.}
	\label{fig:comparsion_ownoffice2}
\end{figure}
\begin{figure}[t]
	\centering
	\includegraphics[scale=0.15]  {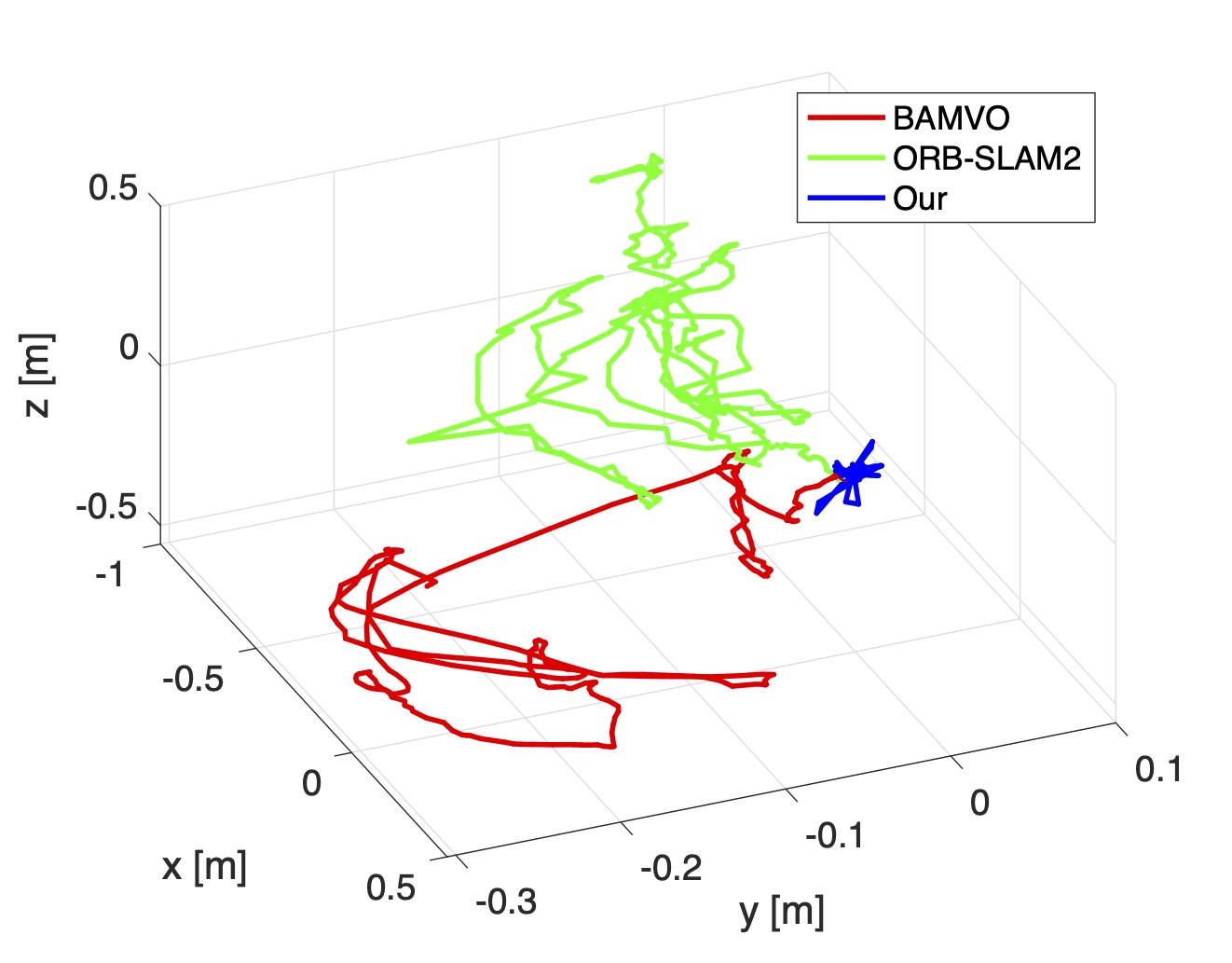}
	\caption{XYZ trajectory results for the environment with a person walking and covering most of the FOV.}
	\label{fig:comparsion_ownoffice3}
\end{figure}

The robustness of DSLAM was further evaluated in three challenging environments with three types of moving objects: a slowly moving object, an object moving with unstructured motion, and a person walking and covering most of the background. In each environment, the ASUS Xtion Pro Live, which was used to record the data, was stationary. Therefore, the trajectory estimated by the methods should be a point. In other words, the distance between the origin and the trajectory is the error. Because DVO SLAM and SPWSLAM do not have open-source code and DVO cannot run in real time on our platform, only ORB-SLAM2 and BAMVO were used for the comparison.

In the environment with a slowly moving object, the reprojected position of the object in the images also moves slowly. Therefore, the projection of the moving object slightly changes in the images of the current and previous frames. The small difference between consecutive frames is a challenge. Therefore, both ORB-SLAM2 and BAMVO are influenced by the moving object. For DSLAM, the front end also cannot determine which points are on the moving object all of the time.
However, the back end can determine static points with data over a longer time span. Because DSLAM is almost uninfluenced by the moving object, the trajectory result remains close to the origin, as shown in Figure \ref{fig:comparsion_ownoffice1}.

In the environment with an object moving with unstructured motion, the object is freely distorted. Distortion can be seen as another challenge because the dynamic points on this moving object have inconsistent and independent motion. Note that the object first enters the FOV. Therefore, ORB-SLAM2 and BAMVO are affected in the first part of the sequence because they cannot remove information disturbed by the appearance of the moving object.
DSLAM can resist the disturbance of the moving object, as shown in Figure \ref{fig:comparsion_ownoffice2}. After the object becomes distorted, the performance of DSLAM is still the best. 
Although the moving object remains in almost the same image region, ORB-SLAM2 and BAMVO still cannot perform well.
In particular, the image region for the moving object cannot be completely excluded by BAMVO.
Therefore, the defective exclusion result influences the estimation output by BAMVO.

The last experiment was carried out in an environment with a person walking and covering most of the FOV.
In this sequence, most of the image is covered by the person and a paper because the person holding the paper is close to the camera. Only a small amount of the information about the static scene can be sensed in most of the sequences. As shown in Figure \ref{fig:comparsion_ownoffice3}, ORB-SLAM2 and BAMVO cannot compensate for the influence of the walking person. Therefore, the trajectories of ORB-SLAM2 and BAMVO are far from the origin. In the results obtained by DSLAM, the dynamic points on both the walking person and handheld paper are discarded from motion estimation. Therefore, the large moving object does not significantly influence DSLAM, and the trajectory obtained by DSLAM remains close to the origin.

\subsection{Analysis of the Efficiency}

In this section, the statistics of the runtimes are presented to evaluate the efficiency of the determination of static points by the front and back ends.
We evaluated the real-time performance of DSLAM on the \textit{walking-static} sequence of the TUM RGB-D benchmark.
In the \textit{walking-static} sequence, the viewpoint of the camera does not change in the sequences since the camera remains almost static at one position. 
Therefore, only walking people influence the runtime of each method,
as summarized in Table \ref{tab:timing_comparesion}.
Compared to ORB-SLAM2, DSLAM performs more efficiently because the dynamic points have been discarded, thereby reducing the computational cost of motion estimation. 
Meanwhile, the standard  deviation of ORB-SLAM2 is larger because its estimation is influenced by the moving objects, as shown in Table \ref{tab:timing_std_comparesion}.
For DVO, the time consumption is unacceptable because the estimation is difficult to complete when people pass by the FOV of the camera. For BAMVO and StaticFusion, their dense operation leads to a poor result even though they already use a downsampled image (320$\times$240 resolution).
Besides, among the feature-based methods (SPWSLAM, ORB-SLAM2, and DSLAM), the methods based on sparse features (SPWSLAM and DSLAM)  have better real-time performance because fewer data need to be processed.
With regard to the runtime of the learning-based techniques, most such methods need additional hardware (e.g., GPUs) and are not designed for real-time applications. 

\begin{table}
	\caption{Comparison of the real-time performance of the motion estimation output in dynamic environments. Only the results for BAMVO and StaticFusion are reported for an image resolution of 320 $\times$ 240 pixels. Here, $\ast$ denotes the running time evaluated in the original papers with more powerful hardware. }
	\centering
	\begin{tabular}{c|ccc@{}}
		\toprule
		 & Mean [ms] \\ \midrule
		DVO &  717.48  \\
		ORB-SLAM2  &    31.34 \\
		\midrule
		Co-Fusion$^\ast$ &83.3 (with GPU)   \\
		MaskFusion$^\ast$  & 200 (with 2 GPUs)  \\
		MID-Fusion$^\ast$  & 400 (with pre-computed data on GPU)\\
		DynaSLAM$^\ast$ & 738.46 (with GPU) \\ 
		\midrule
		StaticFusion$^\ast$ &30 (with GPU) \\
		SPWSLAM$^\ast$ & 22  \\
		 BAMVO  &   67.78 \\
		 Our  &  30.65 \\
		 \bottomrule
	\end{tabular}

	\label{tab:timing_comparesion}
\end{table}

\begin{table}
	\caption{Real-time performance comparison with standard deviation. The results of BAMVO were evaluated at an image resolution of 320 $\times$ 240 pixels. }
	\centering
	\begin{tabular}{ccccc@{}}
		\toprule
		& DVO & ORBSLAM2 & BAMVO & Our \\ \midrule
		Medium [ms] &  109.419   &    31.37     &    66.1960   &   30.4144  \\
		Mean [ms]  &  717.4791   &    31.34   &   67.7775    &  30.6515   \\
		Std. [ms]  &  1467.3502  &   4.8466       &   8.3373     &   3.3191  \\ \bottomrule
	\end{tabular}

	\label{tab:timing_std_comparesion}
\end{table}

The real-time performance for each major step of the static point determination module was measured, and the results are listed in Table \ref{tab:effi}. In the \textit{walking-halfsphere} sequence used for the evaluation, there are moving objects, and the viewpoint of the camera changes over time. Therefore, this sequence is more challenging and suitable for evaluating the efficiency of each step in dynamic environments.
The results in Table \ref{tab:effi} show that the determination of static points by the front end only needs 2.0216 ms on average. This implies that our method has the potential for online applications, and further indicates that DSLAM has the potential for real-time applications.

\begin{table*}
	\centering
	\caption{Real-time performance of the major steps of the determination of static points.}
	\begin{tabular}{@{}llllll@{}}
		\toprule
		\multicolumn{1}{c}{} &  & Module &{Medium} [ms] & Mean [ms] & Std [ms] \\ \midrule
		\multirow{8}{*}{Static Point Determination} & \multirow{4}{*}{Front End} & Build the graph &  1.079 & 1.1172 &  0.3643 \\
		&  & Remove the edges & 0.4665 & 0.4864 & 0.2554  \\
		&  & Separate dynamic group & 0.307 &  0.4180 & 0.3895  \\
		&  & Total & \textbf{1.930} & \textbf{2.0216} & \textbf{0.7609} \\\cmidrule(l){2-6} 
		& \multirow{4}{*}{Back End} & Build the graph & 14.6465  & 14.6582  & 4.6123  \\
		&  & Remove the edges & 95.263 & 91.4375  & 30.5377  \\
		&  & Separate dynamic group & 1.328 & 1.3051 & 0.3327 \\
		&  & Total & \textbf{111.1730}  & \textbf{107.401} & \textbf{35.0098} \\ \bottomrule  
	\end{tabular}
	\label{tab:effi}
\end{table*}

\begin{figure}[t]
	\centering
	\includegraphics[scale=0.2]  {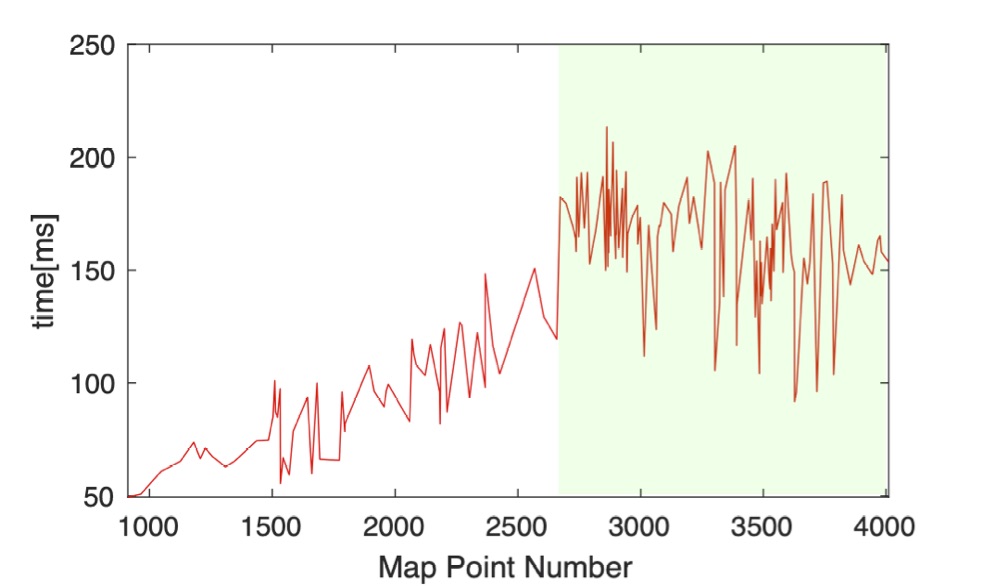}
	\caption{Efficiency of the determination of static points by the back end. The relationship between the number of map points and the time spent by the operation is nearly linear at first. Because the local map points are from a keyframe within the sliding window, the runtime for the determination of the local map points does not grow indefinitely. Moreover, the second half of the graph indicated by the green shading is not smooth because the local map points contain dynamic points that are ignored during optimization.}
	\label{fig:nvt}
\end{figure}

We evaluated the efficiency of the determination of static points in the back end by measuring the time cost with respect to the number of map points. As shown in Figure \ref{fig:nvt}, the relationship between the number of map points and the processing time is nearly linear initially. As the number of map points continues to increase, the runtime does not increase because the number of local map points used for the computation remains stable. Moreover, the number of static points changes when the number of local points is almost constant because there are dynamic points that can be ignored in the calculation. Therefore, the time cost is not smooth.

We note that the static point determination module needs almost 180 ms to optimize 2,500 local map points thanks to the sparse graph construction. The bundle adjustment module requires nearly 200 ms to optimize poses and points and runs on another thread, so the real-time performance of both modules is close. However, because bundle adjustment optimization is needed to determine dynamic points in the local map, it is necessary to complete segmentation as soon as possible before bundle adjustment optimization. Therefore, the efficiency of the static point determination module running on another thread still needs to be improved.


\subsection{Summary of the Results}
The comparison of the accuracy shows that
DSLAM can provide much more accurate results in slightly dynamic environments than the methods that account for dynamic environments. Moreover, DSLAM obtains more competitive results than those obtained with the methods that assume a static world.
In highly dynamic environments, DSLAM obtains more accurate results than most methods considering dynamic environments. The exception is DynaSLAM, which is time consuming to run and needs pretraining.
Moreover, the results shown in the improvement columns in Tables 1, 2, and 3  prove that the robustness of motion estimation is improved significantly in highly dynamic environments because of the proposed segmentation. The comparison with the learning-based methods in Table 4 shows that if a method is only based on learning techniques to filter out moving objects and improve accuracy, it cannot provide the highest performance. Therefore, the proposed method performs better than most learning-based methods. 

The comparison of the robustness demonstrates that DSLAM does not drift away from the ground truth, as do other methods, despite the presence of an unknown moving object with challenging motion in the FOV.

In the analysis of efficiency, although the learning-based methods with geometry cues can provide more accurate results, the computational resources they consume are also enormous. In contrast, the results show that DSLAM has the potential for real-time applications.

\section{Conclusion and Future Work}
\label{sec:conclusion}

In this paper, a segmentation method using point correlations was proposed to divide map points into different components according to their own motion pattern. If there are moving objects, the map points are divided into multiple connected components in which every point has a correlation with the others in that component. Moreover, the proposed segmentation method is not limited to the type of sensor as long as the sensor can provide information related to the point correlations.

Integrating the proposed segmentation method, a SLAM method called DSLAM implemented on RGB-D sensors was proposed to eliminate the influence of moving objects in dynamic environments. 
DSLAM can provide accurate and robust results in dynamic environments.
In the implementation, the sparse construction of a graph with the correlations of adjacent points reduces the computational complexity. An experimental comparison using a benchmark demonstrates that DSLAM can outperform state-of-the-art methods in most dynamic environments. Though it does not outperform DynaSLAM in accuracy, it does not require a GPU and needs less computation time for dynamic object segmentation. Moreover,  the proposed method also could be combined with semantic cues to improve accuracy. We further evaluated DSLAM in three challenging environments. The results demonstrate that DSLAM can provide robust performance in challenging dynamic environments.

In future work, the proposed segmentation method will be extended to sensors such as those in monocular, stereo, and light detection and ranging systems.
The main challenge is that different sensors have different noise models because an accurate noise model plays a crucial role in retaining the available information from the sensor.
Moreover, the performance of the proposed method could be further improved. For example, graph construction can be incrementally implemented to avoid repeated graph creation and reduce complexity. Besides, the special sparsity of the Hessian structure of the point correlation formulation should be used to obtain more efficient solvers. Finally, random sample techniques will be considered to improve the robustness of the estimation.

\ifCLASSOPTIONcaptionsoff
  \newpage
\fi

\bibliographystyle{IEEEtran}
\bibliography{ref}

\begin{thebibliography}{10}
\providecommand{\url}[1]{#1}
\csname url@samestyle\endcsname
\providecommand{\newblock}{\relax}
\providecommand{\bibinfo}[2]{#2}
\providecommand{\BIBentrySTDinterwordspacing}{\spaceskip=0pt\relax}
\providecommand{\BIBentryALTinterwordstretchfactor}{4}
\providecommand{\BIBentryALTinterwordspacing}{\spaceskip=\fontdimen2\font plus
\BIBentryALTinterwordstretchfactor\fontdimen3\font minus
  \fontdimen4\font\relax}
\providecommand{\BIBforeignlanguage}[2]{{%
\expandafter\ifx\csname l@#1\endcsname\relax
\typeout{** WARNING: IEEEtran.bst: No hyphenation pattern has been}%
\typeout{** loaded for the language `#1'. Using the pattern for}%
\typeout{** the default language instead.}%
\else
\language=\csname l@#1\endcsname
\fi
#2}}
\providecommand{\BIBdecl}{\relax}
\BIBdecl

\bibitem{scaramuzza2011visual}
D.~Scaramuzza and F.~Fraundorfer, ``Visual odometry [tutorial],'' \emph{IEEE
  robotics \& automation magazine}, vol.~18, no.~4, pp. 80--92, 2011.

\bibitem{fuentes2015visual}
J.~Fuentes-Pacheco, J.~Ruiz-Ascencio, and J.~M. Rend{\'o}n-Mancha, ``Visual
  simultaneous localization and mapping: a survey,'' \emph{Artificial
  Intelligence Review}, vol.~43, no.~1, pp. 55--81, 2015.

\bibitem{stachniss2016simultaneous}
C.~Stachniss, J.~J. Leonard, and S.~Thrun, ``Simultaneous localization and
  mapping,'' in \emph{Springer Handbook of Robotics}.\hskip 1em plus 0.5em
  minus 0.4em\relax Springer, 2016, pp. 1153--1176.

\bibitem{fischler1981random}
M.~A. Fischler and R.~C. Bolles, ``Random sample consensus: a paradigm for
  model fitting with applications to image analysis and automated
  cartography,'' \emph{Communications of the ACM}, vol.~24, no.~6, pp.
  381--395, 1981.

\bibitem{kerl2013robust}
C.~Kerl, J.~Sturm, and D.~Cremers, ``Robust odometry estimation for rgb-d
  cameras,'' in \emph{Robotics and Automation (ICRA), 2013 IEEE International
  Conference on}.\hskip 1em plus 0.5em minus 0.4em\relax IEEE, 2013, pp.
  3748--3754.

\bibitem{mactavish2015all}
K.~MacTavish and T.~D. Barfoot, ``At all costs: A comparison of robust cost
  functions for camera correspondence outliers,'' in \emph{2015 12th Conference
  on Computer and Robot Vision}.\hskip 1em plus 0.5em minus 0.4em\relax IEEE,
  2015, pp. 62--69.

\bibitem{lee2019robust}
S.~Lee, C.~Y. Son, and H.~J. Kim, ``Robust real-time rgb-d visual odometry in
  dynamic environments via rigid motion model,'' in \emph{2019 IEEE/RSJ
  International Conference on Intelligent Robots and Systems (IROS)}.\hskip 1em
  plus 0.5em minus 0.4em\relax IEEE, 2019, pp. 6891--6898.

\bibitem{palazzolo2019refusion}
E.~Palazzolo, J.~Behley, P.~Lottes, P.~Gigu{\`e}re, and C.~Stachniss,
  ``Refusion: 3d reconstruction in dynamic environments for rgb-d cameras
  exploiting residuals,'' in \emph{2019 IEEE/RSJ International Conference on
  Intelligent Robots and Systems (IROS)}.\hskip 1em plus 0.5em minus
  0.4em\relax IEEE, 2019, pp. 7855--7862.

\bibitem{bescos2018dynslam}
B.~Bescos, J.~M. F{\'a}cil, J.~Civera, and J.~Neira, ``Dynaslam: Tracking,
  mapping, and inpainting in dynamic scenes,'' \emph{IEEE Robotics and
  Automation Letters}, vol.~3, no.~4, pp. 4076--4083, 2018.

\bibitem{runz2018maskfusion}
M.~Runz, M.~Buffier, and L.~Agapito, ``Maskfusion: Real-time recognition,
  tracking and reconstruction of multiple moving objects,'' in \emph{2018 IEEE
  International Symposium on Mixed and Augmented Reality (ISMAR)}.\hskip 1em
  plus 0.5em minus 0.4em\relax IEEE, 2018, pp. 10--20.

\bibitem{chiuso2002structure}
A.~Chiuso, P.~Favaro, H.~Jin, and S.~Soatto, ``Structure from motion causally
  integrated over time,'' \emph{IEEE transactions on pattern analysis and
  machine intelligence}, vol.~24, no.~4, pp. 523--535, 2002.

\bibitem{davison2007monoslam}
A.~J. Davison, I.~D. Reid, N.~D. Molton, and O.~Stasse, ``Monoslam: Real-time
  single camera slam,'' \emph{IEEE transactions on pattern analysis and machine
  intelligence}, vol.~29, no.~6, pp. 1052--1067, 2007.

\bibitem{gutmann1999incremental}
J.-S. Gutmann and K.~Konolige, ``Incremental mapping of large cyclic
  environments.'' in \emph{CIRA}, vol.~99.\hskip 1em plus 0.5em minus
  0.4em\relax Citeseer, 1999, pp. 318--325.

\bibitem{mouragnon2006real}
E.~Mouragnon, M.~Lhuillier, M.~Dhome, F.~Dekeyser, and P.~Sayd, ``Real time
  localization and 3d reconstruction,'' in \emph{Computer Vision and Pattern
  Recognition, 2006 IEEE Computer Society Conference on}, vol.~1.\hskip 1em
  plus 0.5em minus 0.4em\relax IEEE, 2006, pp. 363--370.

\bibitem{klein2007parallel}
G.~Klein and D.~Murray, ``Parallel tracking and mapping for small ar
  workspaces,'' in \emph{Mixed and Augmented Reality, 2007. ISMAR 2007. 6th
  IEEE and ACM International Symposium on}.\hskip 1em plus 0.5em minus
  0.4em\relax IEEE, 2007, pp. 225--234.

\bibitem{strasdat2012visual}
H.~Strasdat, J.~M. Montiel, and A.~J. Davison, ``Visual slam: why filter?''
  \emph{Image and Vision Computing}, vol.~30, no.~2, pp. 65--77, 2012.

\bibitem{endres20133}
F.~Endres, J.~Hess, J.~Sturm, D.~Cremers, and W.~Burgard, ``3-d mapping with an
  rgb-d camera,'' \emph{IEEE transactions on robotics}, vol.~30, no.~1, pp.
  177--187, 2013.

\bibitem{forster2017svo}
C.~Forster, Z.~Zhang, M.~Gassner, M.~Werlberger, and D.~Scaramuzza, ``Svo:
  Semidirect visual odometry for monocular and multicamera systems,''
  \emph{IEEE Transactions on Robotics}, vol.~33, no.~2, pp. 249--265, 2017.

\bibitem{mur2017orb}
R.~Mur-Artal and J.~D. Tard{\'o}s, ``Orb-slam2: An open-source slam system for
  monocular, stereo, and rgb-d cameras,'' \emph{IEEE Transactions on Robotics},
  vol.~33, no.~5, pp. 1255--1262, 2017.

\bibitem{stuhmer2010real}
J.~St{\"u}hmer, S.~Gumhold, and D.~Cremers, ``Real-time dense geometry from a
  handheld camera,'' in \emph{Joint Pattern Recognition Symposium}.\hskip 1em
  plus 0.5em minus 0.4em\relax Springer, 2010, pp. 11--20.

\bibitem{newcombe2011dtam}
R.~A. Newcombe, S.~J. Lovegrove, and A.~J. Davison, ``Dtam: Dense tracking and
  mapping in real-time,'' in \emph{Computer Vision (ICCV), 2011 IEEE
  International Conference on}.\hskip 1em plus 0.5em minus 0.4em\relax IEEE,
  2011, pp. 2320--2327.

\bibitem{steinbrucker2011real}
F.~Steinbr{\"u}cker, J.~Sturm, and D.~Cremers, ``Real-time visual odometry from
  dense rgb-d images,'' in \emph{2011 IEEE international conference on computer
  vision workshops (ICCV Workshops)}.\hskip 1em plus 0.5em minus 0.4em\relax
  IEEE, 2011, pp. 719--722.

\bibitem{cadena2016past}
C.~Cadena, L.~Carlone, H.~Carrillo, Y.~Latif, D.~Scaramuzza, J.~Neira, I.~Reid,
  and J.~J. Leonard, ``Past, present, and future of simultaneous localization
  and mapping: Toward the robust-perception age,'' \emph{IEEE Transactions on
  robotics}, vol.~32, no.~6, pp. 1309--1332, 2016.

\bibitem{saputra2018visual}
M.~R.~U. Saputra, A.~Markham, and N.~Trigoni, ``Visual slam and structure from
  motion in dynamic environments: A survey,'' \emph{ACM Computing Surveys
  (CSUR)}, vol.~51, no.~2, pp. 1--36, 2018.

\bibitem{alcantarilla2012combining}
P.~F. Alcantarilla, J.~J. Yebes, J.~Almaz{\'a}n, and L.~M. Bergasa, ``On
  combining visual slam and dense scene flow to increase the robustness of
  localization and mapping in dynamic environments,'' in \emph{Robotics and
  Automation (ICRA), 2012 IEEE International Conference on}.\hskip 1em plus
  0.5em minus 0.4em\relax IEEE, 2012, pp. 1290--1297.

\bibitem{azartash2014visual}
H.~Azartash, K.-R. Lee, and T.~Q. Nguyen, ``Visual odometry for rgb-d cameras
  for dynamic scenes,'' in \emph{Acoustics, Speech and Signal Processing
  (ICASSP), 2014 IEEE International Conference on}.\hskip 1em plus 0.5em minus
  0.4em\relax IEEE, 2014, pp. 1280--1284.

\bibitem{FlowFusion2020Zhang}
T.~Zhang, H.~Zhang, Y.~Li, Y.~Nakamura, and L.~Zhang, ``Flowfusion: Dynamic
  dense rgb-d slam based on optical flow,'' in \emph{Proceedings of the IEEE
  International Conference on Robotics and Automation}, 2020.

\bibitem{stuckler2013efficient}
J.~St{\"u}ckler and S.~Behnke, ``Efficient dense 3d rigid-body motion
  segmentation in rgb-d video.'' in \emph{BMVC}, 2013.

\bibitem{sun2017improving}
Y.~Sun, M.~Liu, and M.~Q.-H. Meng, ``Improving rgb-d slam in dynamic
  environments: A motion removal approach,'' \emph{Robotics and Autonomous
  Systems}, vol.~89, pp. 110--122, 2017.

\bibitem{li2017rgb}
S.~Li and D.~Lee, ``Rgb-d slam in dynamic environments using static point
  weighting,'' \emph{IEEE Robotics and Automation Letters}, vol.~2, no.~4, pp.
  2263--2270, 2017.

\bibitem{kim2016effective}
D.-H. Kim and J.-H. Kim, ``Effective background model-based rgb-d dense visual
  odometry in a dynamic environment,'' \emph{IEEE Transactions on Robotics},
  vol.~32, no.~6, pp. 1565--1573, 2016.

\bibitem{jaimez2017fast}
M.~Jaimez, C.~Kerl, J.~Gonzalez-Jimenez, and D.~Cremers, ``Fast odometry and
  scene flow from rgb-d cameras based on geometric clustering,'' in \emph{2017
  IEEE International Conference on Robotics and Automation (ICRA)}.\hskip 1em
  plus 0.5em minus 0.4em\relax IEEE, 2017, pp. 3992--3999.

\bibitem{scona2018staticfusion}
R.~Scona, M.~Jaimez, Y.~R. Petillot, M.~Fallon, and D.~Cremers, ``Staticfusion:
  Background reconstruction for dense rgb-d slam in dynamic environments,'' in
  \emph{2018 IEEE International Conference on Robotics and Automation
  (ICRA)}.\hskip 1em plus 0.5em minus 0.4em\relax IEEE, 2018, pp. 1--9.

\bibitem{kim2015visual}
D.-H. Kim, S.-B. Han, and J.-H. Kim, ``Visual odometry algorithm using an rgb-d
  sensor and imu in a highly dynamic environment,'' in \emph{Proc. Int. Conf.
  Robot. Intell. Technol. Appl.}, 2015, pp. 11--26.

\bibitem{huang2020clustervo}
J.~Huang, S.~Yang, T.-J. Mu, and S.-M. Hu, ``Clustervo: Clustering moving
  instances and estimating visual odometry for self and surroundings,'' in
  \emph{Proceedings of the IEEE/CVF Conference on Computer Vision and Pattern
  Recognition}, 2020, pp. 2168--2177.

\bibitem{migliore2009use}
D.~Migliore, R.~Rigamonti, D.~Marzorati, M.~Matteucci, and D.~G. Sorrenti,
  ``Use a single camera for simultaneous localization and mapping with mobile
  object tracking in dynamic environments,'' in \emph{ICRA Workshop on Safe
  navigation in open and dynamic environments: Application to autonomous
  vehicles}, 2009, pp. 12--17.

\bibitem{kundu2009moving}
A.~Kundu, K.~M. Krishna, and J.~Sivaswamy, ``Moving object detection by
  multi-view geometric techniques from a single camera mounted robot,'' in
  \emph{2009 IEEE/RSJ International Conference on Intelligent Robots and
  Systems}.\hskip 1em plus 0.5em minus 0.4em\relax IEEE, 2009, pp. 4306--4312.

\bibitem{tan2013robust}
W.~Tan, H.~Liu, Z.~Dong, G.~Zhang, and H.~Bao, ``Robust monocular slam in
  dynamic environments,'' in \emph{Mixed and Augmented Reality (ISMAR), 2013
  IEEE International Symposium on}.\hskip 1em plus 0.5em minus 0.4em\relax
  IEEE, 2013, pp. 209--218.

\bibitem{zou2012coslam}
D.~Zou and P.~Tan, ``Coslam: Collaborative visual slam in dynamic
  environments,'' \emph{IEEE transactions on pattern analysis and machine
  intelligence}, vol.~35, no.~2, pp. 354--366, 2012.

\bibitem{redmon2016you}
J.~Redmon, S.~Divvala, R.~Girshick, and A.~Farhadi, ``You only look once:
  Unified, real-time object detection,'' in \emph{Proceedings of the IEEE
  conference on computer vision and pattern recognition}, 2016, pp. 779--788.

\bibitem{he2018mask}
K.~He, G.~Gkioxari, P.~Dollar, and R.~Girshick, ``Mask r-cnn,'' \emph{IEEE
  Transactions on Pattern Analysis and Machine Intelligence}, 2018.

\bibitem{dai2016instance}
J.~Dai, K.~He, and J.~Sun, ``Instance-aware semantic segmentation via
  multi-task network cascades,'' in \emph{2016 IEEE Conference on Computer
  Vision and Pattern Recognition (CVPR)}.\hskip 1em plus 0.5em minus
  0.4em\relax IEEE, 2016, pp. 3150--3158.

\bibitem{kitt2010moving}
B.~Kitt, F.~Moosmann, and C.~Stiller, ``Moving on to dynamic environments:
  Visual odometry using feature classification,'' in \emph{Intelligent Robots
  and Systems (IROS), 2010 IEEE/RSJ International Conference on}.\hskip 1em
  plus 0.5em minus 0.4em\relax IEEE, 2010, pp. 5551--5556.

\bibitem{riazuelo2017semantic}
L.~Riazuelo, L.~Montano, and J.~Montiel, ``Semantic visual slam in populated
  environments,'' in \emph{Mobile Robots (ECMR), 2017 European Conference
  on}.\hskip 1em plus 0.5em minus 0.4em\relax IEEE, 2017, pp. 1--7.

\bibitem{barsan2018robust}
I.~A. B{\^a}rsan, P.~Liu, M.~Pollefeys, and A.~Geiger, ``Robust dense mapping
  for large-scale dynamic environments,'' in \emph{2018 IEEE International
  Conference on Robotics and Automation (ICRA)}.\hskip 1em plus 0.5em minus
  0.4em\relax IEEE, 2018, pp. 7510--7517.

\bibitem{he2017mask}
K.~He, G.~Gkioxari, P.~Doll{\'a}r, and R.~Girshick, ``Mask r-cnn,'' in
  \emph{Proceedings of the IEEE international conference on computer vision},
  2017, pp. 2961--2969.

\bibitem{qiu2019tracking}
K.~Qiu, T.~Qin, W.~Gao, and S.~Shen, ``Tracking 3-d motion of dynamic objects
  using monocular visual-inertial sensing,'' \emph{IEEE Transactions on
  Robotics}, 2019.

\bibitem{gordon2018re}
D.~Gordon, A.~Farhadi, and D.~Fox, ``Re$^3$: Re al-time recurrent regression
  networks for visual tracking of generic objects,'' \emph{IEEE Robotics and
  Automation Letters}, vol.~3, no.~2, pp. 788--795, 2018.

\bibitem{yang2019cubeslam}
S.~Yang and S.~Scherer, ``Cubeslam: Monocular 3-d object slam,'' \emph{IEEE
  Transactions on Robotics}, 2019.

\bibitem{strecke2019fusion}
M.~Strecke and J.~Stuckler, ``Em-fusion: Dynamic object-level slam with
  probabilistic data association,'' in \emph{Proceedings of the IEEE
  International Conference on Computer Vision}, 2019, pp. 5865--5874.

\bibitem{runz2017co}
M.~R{\"u}nz and L.~Agapito, ``Co-fusion: Real-time segmentation, tracking and
  fusion of multiple objects,'' in \emph{2017 IEEE International Conference on
  Robotics and Automation (ICRA)}.\hskip 1em plus 0.5em minus 0.4em\relax IEEE,
  2017, pp. 4471--4478.

\bibitem{xu2019mid}
B.~Xu, W.~Li, D.~Tzoumanikas, M.~Bloesch, A.~Davison, and S.~Leutenegger,
  ``Mid-fusion: Octree-based object-level multi-instance dynamic slam,'' in
  \emph{2019 International Conference on Robotics and Automation (ICRA)}.\hskip
  1em plus 0.5em minus 0.4em\relax IEEE, 2019, pp. 5231--5237.

\bibitem{barfoot2017state}
T.~D. Barfoot, \emph{State Estimation for Robotics}.\hskip 1em plus 0.5em minus
  0.4em\relax Cambridge University Press, 2017.

\bibitem{grisetti2011g2o}
G.~Grisetti, R.~K{\"u}mmerle, H.~Strasdat, and K.~Konolige, ``g2o: A general
  framework for (hyper) graph optimization,'' \emph{Tech. Rep.}, 2011.

\bibitem{polok2013incremental}
L.~Polok, V.~Ila, M.~\v{S}olony, and P.~Smr\v{z}, ``Incremental block cholesky
  factorization for nonlinear least squares in robotics.'' in \emph{Robotics:
  Science and Systems}, 2013, pp. 328--336.

\bibitem{barber1996quickhull}
C.~B. Barber, D.~P. Dobkin, and H.~Huhdanpaa, ``The quickhull algorithm for
  convex hulls,'' \emph{ACM Transactions on Mathematical Software (TOMS)},
  vol.~22, no.~4, pp. 469--483, 1996.

\bibitem{engel2017direct}
J.~Engel, V.~Koltun, and D.~Cremers, ``Direct sparse odometry,'' \emph{IEEE
  Transactions on Pattern Analysis and Machine Intelligence}, 2017.

\bibitem{khoshelham2012accuracy}
K.~Khoshelham and S.~O. Elberink, ``Accuracy and resolution of kinect depth
  data for indoor mapping applications,'' \emph{Sensors}, vol.~12, no.~2, pp.
  1437--1454, 2012.

\bibitem{dryanovski2013fast}
I.~Dryanovski, R.~G. Valenti, and J.~Xiao, ``Fast visual odometry and mapping
  from rgb-d data,'' in \emph{Robotics and Automation (ICRA), 2013 IEEE
  International Conference on}.\hskip 1em plus 0.5em minus 0.4em\relax IEEE,
  2013, pp. 2305--2310.

\bibitem{sturm2012benchmark}
J.~Sturm, N.~Engelhard, F.~Endres, W.~Burgard, and D.~Cremers, ``A benchmark
  for the evaluation of rgb-d slam systems,'' in \emph{Intelligent Robots and
  Systems (IROS), 2012 IEEE/RSJ International Conference on}.\hskip 1em plus
  0.5em minus 0.4em\relax IEEE, 2012, pp. 573--580.

\bibitem{kerl2013dense}
C.~Kerl, J.~Sturm, and D.~Cremers, ``Dense visual slam for rgb-d cameras,'' in
  \emph{2013 IEEE/RSJ International Conference on Intelligent Robots and
  Systems}.\hskip 1em plus 0.5em minus 0.4em\relax IEEE, 2013, pp. 2100--2106.

\end{thebibliography}





\balance

\end{document}